# Architecture of a Cortex Inspired Hierarchical Event Recaller

Valentin Puente Varona
Universidad de Cantabria
Santander, Spain
vpuente@unican.es

*Abstract*— This paper proposes a new approach to Machine Learning (ML) that focuses on unsupervised continuous context-dependent learning of complex patterns. Although the proposal is partly inspired by some of the current knowledge about the structural and functional properties of the mammalian brain, we do not claim that biological systems work in an analogous way (nor the opposite). Based on some properties of the cerebellar cortex and adjacent structures, a proposal suitable for practical problems is presented. A synthetic structure capable of identifying and predicting complex temporal series will be defined and experimentally tested. The system relies heavily on prediction to help identify and learn patterns based on previously acquired contextual knowledge. As a proof of concept, the proposed system is shown to be able to learn, identify and predict a remarkably complex temporal series such as human speech, with no prior knowledge. From raw data, without any adaptation in the core algorithm, the system is able to identify certain speech structures from a set of Spanish sentences. Unlike conventional ML, the proposal can learn with a reduced training set. Although the idea can be applied to a constrained problem, such as the detection of unknown vocabulary in a speech, it could be used in more applications, such as vision, or (by incorporating the missing biological periphery) fit into other ML techniques. Given the trivial computational primitives used, a potential hardware implementation will be remarkably frugal. Coincidentally, the proposed model not only conforms to a plausible functional framework for biological systems but may also explain many elusive cognitive phenomena.

## I. INTRODUCTION

Biology, through evolution, has achieved the most flexible, yet efficient, and reliable information processing system known, which is the mammalian brain. Tangible proof of this statement is that the human brain has been able to develop, from scratch, information processing systems as complex as computers. Nevertheless, even with a more reliable substrate, our creations cannot yet compete with biological systems. Although computers are able to outperform biology in narrow tasks, they are far from being able to match biology in terms of flexibility, reliability, and energy efficiency. As in many previous works, this observation is the spark that provokes the question, can we compete with nature by being inspired by its achievements? To explore this possibility, we will focus our attention on the most remarkable part of the mammalian brain: the cortex. The question to be addressed is whether we can develop a practical system with competitive performance in a biologically complex task. If successful, such a system could provide a tool to better understand the intricacies of the biology and/or to build future systems capable of both transcending biological constraints and circumventing foreseeable technological barriers that conventional computers will face in the medium term.

We speculatively hypothesize about how a biological system can Learn, Recognize and Predict (L/R/P/) information. To evaluate such a hypothesis, we will implement a practical system and test its performance on a problem where biology seems to be more flexible than current machine learning methods.

The basic hypotheses are that the cortical columns [1][2]: (1) perform L/R/P/ over temporal signals, and (2) L/R/P/ is maintained by repetition in the input exposure [3]. For a multidimensional continuous complex temporal series, the repetitive sequences (or segments [4]) in the raw data stream are associated with a symbol for each dimension. From the stream of symbols, e.g., a letter and a phoneme, the next level in the hierarchy, will identify repetitions and assign them to a symbol, e.g., a word. This can continue, to higher levels, depending on the complexity of the signal, e.g., sentences, stories, etc… To increase the efficiency of the system, both in terms of resources (i.e., synapses) and energy expenditure, L/R/P/ trades accuracy for temporal and spatial context. Previously acquired knowledge somehow influence the learning of new knowledge. For example, different typographies for a letter should be assigned to the same symbol in the first level. Thus, our speculative hypothesis is that the goal of the biology is to untangle hierarchical events in a complex multidimensional data stream using a highly constrained and brittle set of physical resources organized hierarchically. Not surprisingly complex systems (and complex data) tend to exhibit a hierarchical organization [5].

Neuroscience knowledge evolves at a daunting pace. The introduction of new tools and methodologies provides new insights that, occasionally, require a revision of current understanding. Unlike other sciences, the diversity of models is difficult to reconcile within a coherent theoretical framework. As a result, the amount of available unstructured physical evidence is overwhelming. Without a ground truth, the current path of neuroscience may fail to define coherent functional models [6], or even be detrimental to the ultimate goal of understanding how the brain works [7]. Our approach, given a design objective, is to use certain neuroscience findings as guidance to address specific issues found through the composition of the system architecture (not the other way around). We use some of the currently available knowledge as intuition or guidance for the design decision crossroads. The main sources of inspiration in this work are: [2], which shows the regularity of the cortex and speculatively hypothesizes the existence of a "*Common Cortical Algorithm*", [8]which shows in an animal model that the activity of the cortex depends on the novelty of the stimuli, [9] which describes information segmentation as a fundamental tool for hierarchical organization, [10] which shows how high-level cognitive tasks can be structured using event segmentation, [11] which proposes an alternative model for the activation order of the cortical layers (infra-granular layers before supra-granular layers), [12] which states that the modulation of the learning in the pyramidal cells is controlled by the calcium concentration in the synaptic cleft, [13] which shows the role of hippocampal replay in behaving biological systems, [14] which introduces the cortico-thalamic pathway, and [15] which distills down the current knowledge (and lack of it) about the auditory pathway. Nevertheless, certain design choices made may not be supported by the (current) state of neuroscience (or at least we could not find evidence to support them). They are architected to address lack of conformity between the desired behavior of the system and its observed performance. Among the overwhelming number of possible directions to follow, we always choose a biologically plausible one (from our point of view and with limited knowledge). In many cases, some of the decisions were later confirmed by experimental evidence found in the bibliography.

In addition to not strictly following other neuroscience computational models, this work will not use any mathematical tools. In fields such as computer architecture, such approaches are of limited use. It is not possible to model such a complex system in sufficient detail to accurately predict system behavior. Mathematical formalism may be useful in certain design phases and components, such as validating memory models or early design of cache coherence protocols but are not appropriate in many others. Therefore, although a computer is a dynamical system, formal tools successfully used in experimental sciences to model physical and dynamical systems, have a limited applicability.

Abstraction and layering are the cornerstones for designing or understanding a computer system. From there, simulation, over many levels, is the fundamental exploration tool. Final proposals are usually the result of an understanding of the internal state of the system. Often, since we cannot have all the evidence given that the software stack is usually complex and diverse, innovations are guided by experience and intuition. Above all, any proposal should be feasible within the technological constraints in use. In this paper, we have taken a similar approach. First, we speculatively hypothesize that biology may also use some form of abstraction to cope with the brittleness of the physical substrate. We choose one particular layer in such an *unknown* hierarchy of abstractions. We will tackle a relatively simple problem, from a biological point of view, under biologically plausible conditions. We focus on providing a simple but functional design. The writing style, aligned with a typical computer architecture paper, describes the proposal to the reader and evaluates its performance in the task under consideration. As a final objective of this process, we arrive at an architecture, denoted *Hierarchical Event Recaller* (HER). However, it should be understood as an algorithmic construct. In each description we will include speculative hypotheses on how biology might approach each problem. Such a combination of conjectures could be a potential candidate for the long time hypothesized [2][16] (but controversial [17]) *Common Cortical Algorithm*. Despite being of inspired mostly by neuroscientific findings, the proposal offers some plausible explanations for many elusive cognitive-level effects, which weakly suggests that it may contain a sliver of the ultimate goal.

II. CORE PRINCIPLES OF HER

A. **Temporal Prediction based on Sequence Segmentation**

The central element, used in different modules, is a biologically *Sequence Memory* proposed in [18]. The core idea is to identify and label segments (or sequences) in its input stream at each level of the hierarchy. Each identification is forwarded to the next level (hereafter referred to as rungs[1]) in the hierarchy. Each rung will L/R/P sequences with higher semantic meaning. By segmenting the input stream in each rung, it will be possible to handle very complex patterns with a limited set of resources (i.e., cells and synapses). The current input and the previous (or internal or hidden) state of each *Sequence Memory* determine the current state. Note that the *previous* instant is further away in physical time as we progress through the hierarchy. Nesting sequence segmentation, by integrating at different timescales across the hierarchy, it will be feasible to manage complex sequences cohesively. Note that this is similar to how the activity can be integrated and structured in biological systems [10][19].

---

[1] To avoid confusion between the levels of the cortical column and the hierarchy, the latter is hereinafter referred to as the "rung".

## B. Short and Long Term Learning Modulations

To ensure flexibility, learning is stochastic. The probability of learning is determined by the degree of *knowledge* of the current input sequence. If the most recent values of the sequence were incorrectly predicted (i.e., the sequence is *unknown* in the short-term), the probability of changing a synaptic permanence is significant; otherwise, it will be very low. Each group of *Sequence Memories* in the cortical column (CC) will have two states, according to the input activity seen in a few time steps back: *Known* and *Unknown*. Similarly, over a longer period, the entire CC will have equivalent states. Learning can be enhanced or attenuated as needed, depending on the CC's long-term familiarity with the input. In the first state the short-term learning modulation is increased in all layers of the CC. In the second state, the short-term learning modulation is reduced to the point where learning is turned off in certain layers.

## C. Input Dimensionality Reduction, Pattern Disambiguation, and Feedback

Each *Sequence Memory* handles a limited number of different values. The dimensionality of the input to each *Sequence Memory* is reduced using a k-winners take all inhibition. By using a network with many silent synapses, we will project each input value into a few clustered representations (each cluster, formed by a group of similar sparse distributed representations, in our terms, associated with a **symbol**). Sequence memories can accurately track the temporal context of each symbol (using a state representation based on combinatorics).

Replication can be used to increase system robustness or to facilitate disambiguation of complex patterns. Multiple inhibition processes using a random sampling of the same input, using silent synapses.

Since the system is "flexible" and continuously learning, changes in the representations of the symbols are expected. The main purpose of the intracortical feedback is to help to stabilize the inhibition processes (i.e., to stabilize symbol generation). When multiple potential winners in an inhibition process have equal overlap with the input, the feedback facilitates a consistent outcome playing the role of tie-breaker. Because ties are frequent, feedback has a pronounced effect on system stability despite small local effects. If necessary, feedback-like signals can be generated "externally" to perform supervised symbol formation (not explored in this work).

## D. Spurious Identification Filtering, Learning Acceleration, and Learning Gating

In addition, during the learning process an auxiliary *Sequence Memory*, which may resemble a part of the functionality of the *hippocampus*, oversees the prevention of contamination of higher rungs of the hierarchy with spurious or incomplete identifications. This module will be necessary in all the rungs of the hierarchy. They can be assigned statically or allocated on-demand from a centralized structure. The module is assigned when the long-term knowledge status of some CC in the next rung is *unknown*. If the module is unfamiliar with the current data, learning in the next rung CC is deactivated. Once all CCs in the next rung are familiar with the data, the component is deallocated. While the component is familiar with the data and a certain CC in the next rung is unfamiliar, the predictions are used recursively to generate predicted sequences that help to stabilize the next rung CC input stage.

## E. Lateral Contextualization

Each CC will have two types of inputs: (1) vertical inputs coming from a single preceding cortical column $CC_{prev}$ used to form a temporal reference frame for the sequence, and (2) lateral inputs coming from neighboring columns to a CC used to uniquely identify the sequence in the spatial context of $CC_{prev}$. Compounding vertical and horizontal input, within the reference frame built by the vertical connections, a CC will generate a single identification modulated by the sequence perceived by a neighboring CC.

## F. Attention and Speculative Pattern Identification

Each CC will have the ability to speculatively predict the input sequence identification of higher rungs of the hierarchy. If the expectations of the next rungs in the hierarchy (i.e., the provided feedback) match some form of higher order prediction in the current rung, we can forward a prediction of the sequence without waiting to receive it completely. This will speed up the identification process and adjunct dense projection in collapsing similar sequences into a common identification. Corticothalamic loops inspire this idea.

## G. Supervised learning (and output stabilization)

The system can incorporate mechanisms to direct the learning output of each CC to specific objectives, both in timing and representation. This can also be used to stabilize the CC output if it is in a long-term known state. Although such learning/identification targets can be self-generated, the proposed system lacks the mechanism to do so. It will be necessary to incorporate action capabilities (e.g., to change the sensors) and/or certain reward-based selection policies. This idea is inspired by the ascending pathway of Corticostriatal loops.

### H. *Brainstem and Context Stabilized Encoding.*

The peripheral circuitry will reduce as much as possible the precision of the information supplied to the brainstem, while not degrading the cortical disambiguation capabilities. Input variability on sensory data will not be transferred to the cortex. To achieve this, we use the same dimensionality reduction described in point II.C. Similarly to intracortical feedback, to increase stability, the brainstem will use cortical context to modulate the inhibition process. Given the physical constraints, the context is built on the cortical output signal using densely connected dendrites.

### III. ELEMENTS OF THE CORTICAL COLUMN

Inspired by the morphology of the biology system, we divide the Cortical Column into the six components presented in Figure 1. Next, we will proceed to briefly describe each one.

### A. *Input Projectors(L4) and Projection Trackers (L23)*

Both components form the backbone of the cortical column. They are composed of a variable number of replicas, that individually resemble the HTM/CLA[20] components, (i.e., *Spatial Pooler*[2] in L4 and *Temporal Memory* in L23). The **Input Projector** (L4) is divided into a set of individual replicas (ten in the figure) that according to the connectivity between the input and the output bits, carry out a *k-winners-take-all* (k-WTA) inhibition [21]. The inputs and outputs use Sparse Distributed Representation (SDR)[22][23]. A value is encoded as a long vector of bits where only a few of them (~2-3%) are set. Each L4 replica has full synaptic input-to-output connectivity, most of the synapses being silent (i.e., with permanence below the threshold of connection). Certain connected synapses (i.e., with permanence above the threshold) and input determine the projection into the output. The input of each L4 replica is composed either of a lateral input or a subset of the forward input. The forward input is originated by the preceding CC (or a part of the sensor in the sensorial cortices). Lateral input comes from neighbors CC of the preceding CC (or neighboring sensors). Lateral input is required for input binding. The Cortex Architecture (see Section V) will detail how lateral input is modeled in HER.

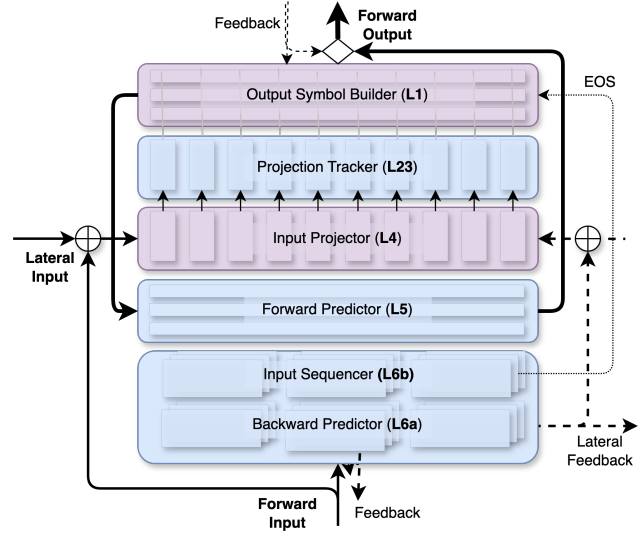

**Figure 1 High Level Description of the Components of the Cortical Column**

L4 performs a Locality-Sensitive Hashing (LSH) of the input, i.e., it will provide a random locality sensitive "hash" of it [24]. Therefore, certain deviations in the input value will be projected on a sparse value with small variations from trial to trial. We denote this group of close SDR as "Symbol". The range of such variations is dependent on the connectivity and the previous experience. To push apart different symbols, silent synapses that reach a permeance value of zero are pruned out (i.e., cannot be formed again). From outside, the L4 functionality can be understood as some form of low-dimensional manifolds, but ultimately its task is to reduce and separate as much as possible the symbols perceived by L23. Although this is consistent with experimental evidence [25], our interpretation is far from the current computational neuroscience approach [26] (more in line with physical dynamical modelling tools).

Learning in L4 is required to filter out noise (e.g., due to cross activity in other CCs, input variations, etc,…). Also minimizes representational drift (e.g., due to transitory learning in the previous CC) progress toward L23. Hence L4 need to keep the cluster of the symbol as compact as possible. A densely connected L4 will tend to cluster similar inputs in a stable output. When different inputs have some shared dimensions, depending on the initial connectivity, the output might collapse in the same output, hindering system disambiguation capabilities. To prevent this, it may be necessary to use multiple replicas. At boot time, once the system is exposed to the sensory input, L4 synapses permanence tends to stabilize. Later, unused synapses can be pruned away with no impact on system behavior. Usually, L4 will map a range of inputs onto a few different values (less than 10). Those values once learned will condition new sensory data, which may explain cognitive level effects such as [25], where learning new patterns may be constrained by the already acquired knowledge.

---

[2] Our interpretation of the functionality of L4 diverges from [20] because it ignores the inherent limitations of the temporal memory to handle a reduced number of symbols with maximal distance symbols.

Each replica in the *Sequence Projector* feeds an individual replica in the **Projection Tracker** (L23). L23 replicas are narrow *Sequence Memories*. They are composed of a set of cells (inspired in the apical tuft of a pyramidal cell). Each cell will have a group of distal synaptic segments (DSS), called branches. The number of branches determines the number of contexts where the owner cell might be part of ensemble. Each cell is connected to one input coming from the sequence projector. The sequence projector output can be envisioned as a proximal dendritic segment. L23 will learn and predict each projection of the input sequence coming from the corresponding L4 replica.

The working principles, inspired by the NMDA dendritic spikes [27], are focused on predicting the next active L23 replica input. Such prediction is carried out using the current input and state of the element. In each input bit, branches of dendritic segments are used to encode this state. Each branch will be in predictive state when the number of activations in synapses connected to other branches in some dendritic segment is above a certain threshold. When a branch is in predictive state, the corresponding bit is set as predicted for the next time step. A group is active (and hence can contribute to the prediction of others in the next step) when being in predictive state, the corresponding input bit is set (i.e., was correctly predicted). Our terminology (dendritic segment, branch, cell) departs from the HTM (dendritic segment, cell, column) [20] because it seems closer to biological systems: dynamic compartmentalization in neurons enables branch-specific learning [28]. Therefore, branches could be crucial to produce contextualized predictions.

We speculatively hypothesize that biological systems use the same approach. A large number of mini-columns sensitive to the similar input features, which is well known in the case of V1[29]. This might be useful to reduce the synaptic load of L23 under the natural stochastic input activations or input similarity. This projection idea might be the cornerstone to build invariant representations of the input across the cortex, since we are transforming any new input into a representation that depend on innate connectivity and the previous experience.

Experimentally, it is observed that using a lower connectivity for L4 in the first rung, the diversity of symbols, and hence sequences identified by L23, is noticeably large. This breaks the rationale of hierarchical storage since the first rung in the hierarchy is identifying a wide range of long sequences. This also renders the system too fragile to cope with any input variation. In a way we trade instantaneous precision for temporal correlation, where *Sequence Memory* excels. While all the groups of replicas of L23 state change independently according to the knowledge of the projection provided by each L4 replica, they operate in a coordinated way. The recent average number of correct input bit predictions determines the *Global State*. L23 state regulates the short-term probability of learning at synaptic level for the whole component.

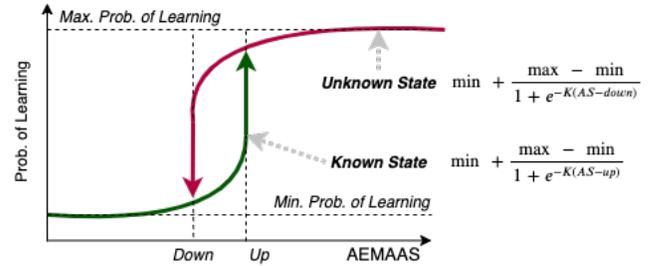

**Figure 2 Temporal Memory Novelty-Based Learning**

This probability is computed as a function of the *Global State* and the Average of the Exponential Moving Averages of the Anomaly Scores of all replicas (AEMAAS$_{L23}$), where Anomaly Score is defined as the ratio of the number of correctly predicted bits and the number of active bits in the output. The Global State is set to **Known**, when L23 is *familiar* with the input sequence, or **Unknown**, when L23 is *not familiar* with the L4 projections[3]. The change from one state to the other is controlled by AEMAAS$_{L23}$ or AS and two parameters, called *Down* and *Up*. As depicted in Figure 2, in each state the probability follows a logistic function centered in the corresponding threshold, to form the hysteresis loop. The motivation for this design is to prevent the learning of an *unknown* value unless it belongs to an *unknown* sequence. Spurious input values in a known sequence will likely be ignored. Additionally, it will allow the collapse of similar sequences in the same final representation. The modulation is applied not only to learning but also forgetting, otherwise noisy inputs can inadvertently decay some synapses. The alpha value in the exponential moving average serves as a knob to control how relevant the past is (i.e., a small value like 0.01 means the past is relevant, while 0.9 means the past is almost unimportant). In combination with this parameter, the threshold parameters *Up* and *Down* govern how the network reacts to new information. *Up* is interpreted as the proportion of bits that can change

---

[3] Although the idea is loosely inspired in UP and DOWN neuronal state [35], we are unable to find a clear link between the experimental evidence and *known-unknown* functional definition.

in recent predictions (according to alfa) of a subsequence before changing to the next.

The constant *K* determines the steepness of the function. Since the rationale of the learning hysteresis cycle is to prevent learning if the system is familiar with the input. When the Sequence Memory (SM) is in a *known* state, the probability of learning should be negligible. Therefore, *K* could be significant (~1000), although it could vary depending on the state of the cortical column.

The main reason for this design is to "segment" the sensory stream. Even when the system is exposed to complex data, it will naturally predict "subsequences" (because the hysteresis cycle prevents all the data from being combined into one continuous prediction). Similarly, when the system is exposed to new data, the interference with the already known data is minimized. The length of such "segments" (and the required synaptic resources) depends on the previously introduced parameters: *alfa*, *Up*, *Down* and *K*. In the case of L23, to improve system stability we need to look quite far in time, using alfa ~0.01, Up ~0.1 and Down ~0.01.

This idea is based on the intuition that novelty promotes learning, which is a well-known experimental observation [30]. We speculatively hypothesize that in biological systems the learning properties in each level of the cortical column is determined by the extracellular Calcium concentration ($[Ca^{2+}]_0$) in the synaptic cleft. This concentration determines a global state for all the pyramidal cells affected. The long-term potentiation (LTP) and depotentiation (LTD) are influenced by this variable [12]. If an input sequence is correctly predicted, i.e. a dendritic spike occurs, the calcium influx in the apical dendrites will be higher [31] and consequently $[Ca^{2+}]_0$ altered [32]. Since $[Ca^{2+}]_0$ mediates the astrocytic response in their processes [33], our hypothesis is that the glia will react to such changes providing a cortical column level response [34]. This is consistent with the fact that processes of a single astrocyte domain comprise many neurons. Note that gap junctions between astrocytes can give rise to a syncytium, which is capable of affecting hundreds of thousands of synapses in a coordinated manner [35]. Although the exact role of astrocytes in learning it is not fully understood [36], their critical role is well established [37]. While the slow somatic change in calcium concentration for astrocytes is not compatible with electrochemical rate of change in neurons, astrocyte small domain changes seem to be much faster. In any case, our hypothesis, is that processes effects on the synaptic cleft are where the key effect of the astrocytes arises. Although we are not aware of experimental evidence in the literature for $[Ca^{2+}]_0$, recent observations, such as [38][39], point in that direction.

This work provided insights about how neuronal activity affects peripheral astrocyte processes (PAP), and how glutamatergic activity impacts on the concentration of $K^+$ and neurotransmitters in the synaptic cleft. Such a scenario will depolarize the astrocyte membrane, potentially activating voltage gated calcium channels (VGCC). This will increase the $[Ca^{2+}]_0$ influx into the astrocyte, reducing the learning at the postsynaptic cell. If the influx is large (e.g., due the low permanence of the synapse because the association is still not learnt), this will significantly increase the $[Ca^{2+}]_0$ influx to the point of activating the IP3 pathway (via G-protein), depleting the $Ca^{2+}$ reservoirs. This will considerably increase the $Ca^{2+}$ in the PAP, might rise the $[Ca^{2+}]_0$ to LTP compatible levels [40]. Thus, a certain degree of coupling between the glutamatergic activity and astrocytic $Ca^{2+}$ could be observed [41]. The hysteresis required by our model could be supported by the IP3 pathway, which uses a secondary messenger, expressed as a certain reluctance to re-enable learning ($k \rightarrow u$ state change). As predicted by the model the PAP $[Ca^{2+}]_0$ oscillations are consistent with neural activity [38]. The intuition of using logistic curves in the hysteresis cycle is weakly supported by the experimental evidence that astrocyte calcium responses to sensory activity, as with most single cell or population responsiveness to electrical stimulation, fit a sigmoid function [42]. In vivo studies such as [43], have extended this observation by showing a hysteresis pattern between sensory-induced neuronal activity and astrocyte intracellular calcium, as required by our model. Note that although our model refers to "knowledge" as being dependent on PAP calcium, this class of experimental evidence can be a first-order proxy for it.

To improve system behavior, L4 must learn as fast as possible. There is no *short-term* modulation. In contrast to L23, doing so in L4 delays symbol stabilization, which is counterproductive for L23 synaptic loading. Compartmentalization of learning in pyramidal cells [44] weakly supports the biological plausibility of this design decision.

In our design, plasticity-induced changes in L4 could increase learning interference. Synaptic pruning reduces the chances because an input-output relation never experienced during the learning, will not contribute to outputs used in already known symbols. The representation of new symbols will most likely be accommodated in other outputs. However, during the steady state, the inherent stochastic effects can create internal noise. To prevent that noise from altering the acquired knowledge, L4 learning is disabled at some point. The *Long-term* modulation is used to autonomously decide when the system is "trained" (i.e., no long-term modulation means L4 learning should be disabled). Intuitively if all the CC in a rung of the cortex have a low anomaly score in the long term (from hundreds of milliseconds to several seconds), L4 learning is turned off. If the long-term modulation is restarted, by a new input stream, learning will be reactivated.

Synapses between inactive inputs and inactive outputs are not affected by learning. They can be pruned proactively. However, if pruning occurs too early, the system may converge the new input sequence to an already known symbol. If pruning is late, there is little chance of having interference with prior knowledge, since the outputs that can be affected by the new inputs cannot conflict with it. A hypothetical mechanism for judging when the CC is familiar with the input might open the possibility of doing this. Fortunately, we have one: see Section IV.D.

Here (and in forthcoming modules) we assume a synchronous operation (in the sense that all input/output bits are active simultaneously at a given "cycle"). While this seems non biologically plausible, the existence of cell assemblies in its firing patterns is hypothesized to be the fundamental unit of neural syntax [45]. We speculatively conjecture that axonal junction gaps might support such a mode. Anatomical and physiological evidence suggest it [46]. Hence, all pyramidal cells that are ready to fire (i.e., resulting winners of the inhibition) might be encouraged to do so near-simultaneously when one of them fires. Note that, in contrast with chemical channels, junction gaps allow sub-millisecond communication between cells. Therefore, if the percolation threshold [47] is reached by a group of cells (such as a CC), it could be possible to achieve a pseudo-synchronous firing of all the winners in the module.

Finally, sequence segmentation based on the $k \rightarrow u$ transition could be supported by the distinctive neuronal activity under boundary conditions. This event segmentation has already been observed in the hippocampus [9][30]. Perhaps not coincidentally, at the cognitive level, prediction errors create sequence or event boundaries [48][49][50]. We speculatively hypothesize that such a phenomenon is pervasive across all levels of the cortical column and rungs of the cortical hierarchy (not just at cognitive levels and/or the hippocampal complex) and it is the key element in achieving perceptual sensory segmentation. By preventing chaining of predictions across sequence boundaries, it facilitates hierarchical model storage.

B. *Backward Predictor(L6a) and Input Sequencer (L6b)*

Similarly to *L23*, *L6a* is responsible for predicting the next input in the forward port of the cortical column (CC). In addition, if the forward input of the CC is at the boundary of a sequence *L6b* will start the mechanism required to build and forward an identification of the sequence to the CC in the next rung. Unlike *L23*, to have a useful prediction of preceding modules (CC or sensors), the input is not a projection (i.e., there is no inhibition involved). The same forward input will reach both L6a and *L6b* one cycle ahead of *L23*, which is inspired by [11].

Although the same input is used, the dual functionality of the layer cannot be combined. Hence, two Temporal Memories are used. **Input Sequencer** (*L6b*) , inspired by the Subgriseal layer [51], is responsible for finding short-term regularities in the input stream. To take advantage of hierarchical organization we need to form subsequences of relatively short length (e.g., length of ~4-10 inputs). Learning parameters are tuned for fast response to unexpected changes in the input (i.e., consider a very short term of the input sequence to determine the $AEMAAS_{L6b}$, typically an α parameter in EMA close to 1). Such fine segmentation of the input will degrade the feedback availability (e.g., up to 75%). In the first term, to alleviate such a situation, we duplicate the layer as L6a (or **Backward Predictor**) with learning parameters tuned to have a much softer response to anomalies (i.e., α parameter in EMA close to 0) and dedicated exclusively to producing backward predictions (i.e., feedback). In the second term, this layer is also used to modulate the short-term learning curves of the remaining temporal memories of the CC. The layer split was inspired by strong inhibitory connectivity between *L6b* and *L1* [52] (i.e., where sequence segmentation is supposed to occur). In addition, only *L6a* projects back to the thalamus and *L4*[53][14] (i.e., supposedly feedback).

Since the receptive field of *L6a/b* in biological systems is wider than of L4 [54], each replica of *L6a/b* could receive multiple groups of combined input flows[4] of the *L4* replica groups affected by the forward input. This is equivalent to combine the different projections (of the same data) from the previous CC or sensors into a wider SDR. Thus, the predictions are supported by a longer number of synapses (because the number of active cells in the previous time step is larger[20]). This increases the stability of the prediction. Since temporal accuracy is paramount to system stability, only the *L6b* receptive field can be increased (ameliorating the synaptic load in *L6a*). To compensate stochastic learning, synaptic transmission noise [55] and representational drift, each Sequence Memory of *L6a/b* might be replicated. Since each input flow will have multiple predictions available, *L6a* will provide more informative feedback (according to the number of predictions in each bit). This seems to be aligned with the fact that *L6a* cortico-thalamic (CT) axons are metabotropic [56] (i.e., the post-synaptic effect of the spike is weak but additive over time). Typically, the system is configured in such a way that the number of *L23/L4* replicas is equal to the replicas of *L6a/b* divided by the number input flows. This is suggested by the cytoarchitecture of the biological mini-column [2].

---

[4] Although forward input comes from the previous Cortical Column/sensors, the input can be composed of multiple sequences of different projections, denoted input flows (see next sub-section).

A design choice of *L6a/b* is that it will not learn if *L23* is in *Unknown* state. The rationale of this decision is to prevent the learning of unstable forward inputs before *L23* discovering stable sequences. Since *L4* projections are clustered, *L23* finds sequence repetitions more accurately than *L6a/b*. By coordinating the learning, we can easily transfer this awareness to *L6a/b*.

This may be biologically plausible. On the one hand, a distinctive characteristic sub-granular pyramidal cell, is the extensive presence of hyperpolarization-activated cyclic nucleotide–gated channels (HCN) [57]. The activation of such channels induces inhibition of the apical dendrites via the hyperpolarization activated current $I_h$ [58] [59]. A likely hypothesis is that such channels are used to disable the learning in the infra-granular cells apical tree in accordance with the state of the supra-granular layers. The activation of HCN happens when the $[Ca^{2+}]_0$ is maximal and is meditated by astrocytes via acetylcholine segregation coming from the cholinergic system [60]. On the other hand, the laminar organization of astrocytes [61] might suggest an alternative approach: certain astrocytes which map his processes over L6a/b synapses react to L23 knowledge degree.

The learning probability curve of L6a/b replicas uses their own average anomaly score (AEMAAS$_{L6a/b}$). Like in L23, LTP/LTD will be performed in the *L6a/b* following the rules depicted in Figure 2. For *L6b*, at the last point in the sequence (i.e., AEMAAS$_{L6b}$ is above the *Up* threshold), the synaptic state is reset. *L6a* will generate feedback to the previous CC. See Section V.C. for a description of how the feedback is interpreted. In addition, the mere presence of *L6a* feedback can be used as an indicator of how familiar the column is with the input data. Section IV.A describes how this is used in conjunction with a hippocampal construct to avoid polluting the hierarchy with noise induced by transient learning, boosting, and deactivating learning.

Although the states of *L23*, *L6a* and *L6b* are independent, which is consistent with the fact that the laminar structure of the glia is not the same as the excitatory cells [61][62][63], the input forward temporal rhythmicity is used to build sequence representations to the next CC. No interaction is required from *L6b* to *L6a* or *L23* at the end of sequence (EOS). Active dendrites in L6b at EOS event should be adjusted accordingly.

Given the smaller time constants of *L6b*, its synaptic load is expected to be lower than *L23* and *L6a*. This may be consistent with the fact that L6b is so small that it is barely distinguishable in many species of biological systems [51]. Indeed, L6b is parameterized for rather short sequences, which also makes sense for a relatively high LTD, since multiple contexts can be segregated into different subsequences.

### C. *Output Symbol Builder (L1)*

This component is responsible for assigning the output "symbol" of the CC, for a given input sequence (according to *L6b* state changes). Therefore, L1 will perform the dimension reduction (i.e., the multiple inputs belonging to each sequence will be converted into a single SDR). The size of the intersection of the output SDR of two sequences should be proportional to the degree of similarity. To do this, we use the correctly predicted cells in the *L23* replicas. Once the input forward sequence boundary is detected by *L6b*, *L1* performs an inhibition process (k-WTA), using as inputs the dendrite groups correctly predicted by all *L23* replicas in the last point of the sequence. Note that *L6* input is one-value ahead of *L4*. This design choice is inspired by the cognitive "peak-end rule" [64]. Using mid-sequence predictions is less biologically plausible, especially in the upper rungs of the hierarchy, because it requires tracking predicted cells that may be very far back in physical time. Additionally, this decreases the difficulty of separating similar sequences, since we attend only to one predicted value per sequence (the last one). The result of the inhibition is modulated by the feedback provided by the next CCs in the hierarchy, by slightly increasing the overlap of the expected output bits.

The k-WTA result depends on the initial connectivity of *L1*, the current cells correctly predicted by *L23*, and the next CC *L6a* expectations (*L6a+*). As in *L4*, *L1* is densely connected to reduce the similarity between symbols. As in *L4*, multiple replicas for *L1* can be used to prevent output collapse. Each *L1* replica, is connected to a subset of the *L23* replicas associated with each input flow. The idea is to reduce the number of symbols per replica and increase the capability of representation just by combining more replicas. Since *L1* and *L4* have similar purposes, they both use the same learning rules.

The expectations of the cortical column in the next rung (i.e., the feedback) can be used to support the output stabilization of the LSH performed by *L1*. Thus, a *L6a* functional objective is to stabilize the clustering performed by *L1* in the preceding CC. A more detailed discussion of feedback as a tiebreaker in inhibition is provided in Section V.C.

Unlike *L23 or L6a*, the state of *L6b* is predominantly influenced by the most recent events (i.e., the alfa value in the EMA of the AS is close to 1) and the *Up* and *Down* parameters are set to 0.1 and 0.5. Therefore, *L23* sequences will typically be longer than *L6b* sequences. The rationale for this choice is to allow a temporal context to be built in *L23* to generate the output symbols in *L1*. Intuitively, *L23* builds in advance the sequence of input symbols that the upcoming cortical column *L6b* will generate with output symbols. *L6a* is more like the parameters of *L23*, so it will also predict long sequences. Astrocytic laminar specialization [63] may support the plausibility of such heterogeneity between CC layers.

We could not find clear evidence for the EOS mechanism (and subsequent symbol generation). Since it must be a fast phenomenon, it is most likely not mediated by astrocyte dynamics, but by some kind of fast inhibition. A weak indication may be the existence of a class of inhibitory cells in L1 that project back to subgranular layers [65][66] and that the L6b neurons strongly innervate cortical L1[52]. These neurons are influenced by the cortical column input and the matrix of the thalamus, which, according to our model (see Section V.D), could be involved in the external generation of an EOS via the cortico-striatal loop [67][68][69]. Other locally projecting *L1* interneurons may also be involved in the symbol formation (i.e., sensing the predicted branches in the firing *L23* cells and performing the k-WTA process) or helping to stabilize the output representation when the CC is *known* state. The functional relevance of *L1* neurons is still in the early stages of understanding and has been referred to as the "crowning mystery" [70]. The final piece of evidence supporting our model is the ability to somehow sense the dendritic spikes in *L23* cells to generate the output of the cortical column. We could find no clear evidence for such a structure in the current neuroscience literature.

### D. *Forward Predictor(L5) and the Cortico-thalamic Loop*

Predicting the symbol before receiving the full sequence ($S_0$) will both increase cortical resilience to input variability and improve identification speed. For this purpose, a *Forward Predictor* (*L5*) is added to the CC. *L5* will learn the sequence generated by *L1* replicas (denoted as $S_+$) toward the next rung. *L5* uses several *Sequence Memories*, each one learning a horizontal subset of the values of $S_+$. Additionally, to improve predictions accuracy, each Sequence Memory can be replicated multiple times. Therefore, *L5* is replicating the work of *L6b* in the next CC (*L6b+*). *L5* will learn only if *L6b+* is learning (Section IV will introduce a mechanism to locally identify this situation). As with any *Sequence Memory*, the learning rate and sequence boundaries are determined by the novelty of the input using the mechanism presented in subsection A). As in

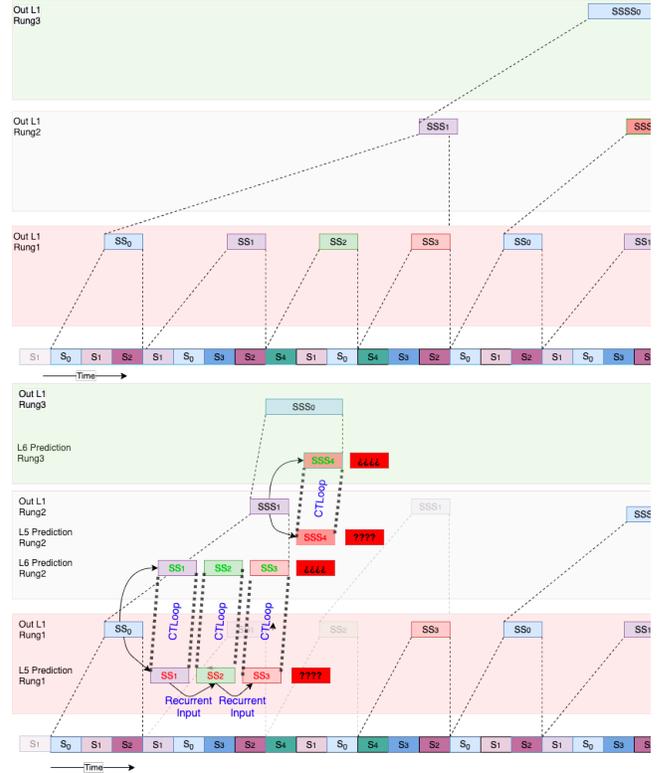

Figure 3 Accelerating high-order identification: (above) Sequential approach, (below) Cascading predictions using the Cortico-thalamic loop (CTLoop). Assuming only one CC per rung.

*L23, L6a/b*, learning hysteresis in *L5* prevents long sequence predictions. To enhance prediction accuracy, L5 will utilize learning parameters that fall between those of L6a and L6b.

Anticipating the current sequence symbol is not enough. We need to somehow quantify how reliable such a prediction is before we can use it. The expectation of the next cortical column (i.e., *L6b+*) is used to estimate how accurate it is. Thus, if *L5* and *L6b+* predictions agree, we can consider that the prediction can be speculatively forwarded (i.e., sent without waiting for the reception of the whole sequence). Even more, by recurrently feeding *L5* with its own prediction, we can anticipate multiple sequences in the current rung. By cascading predictions, we can speculatively perform pattern matching in upper rungs of the hierarchy at an early stage of the sensory stream. Remarkably, the entire cortex may respond requiring only a fraction of the input.

An example of how identification is accelerated up the hierarchy is shown in Figure 3. In the bottom rung we receive a sequence of sensory information in 21-time steps of 5 symbols $\{S_0,..,S_4\}$. From this, the first rung identifies four different sequences (previously learned). The input data is arranged in 6 subsequences, labeled as $\{SS_0,..,SS_3\}$ by the *L1* of the 1st rung. From them, second rung identifies two sequences, labeled as $\{SSS_0, SSS_4\}$. The last rung identifies from them a single sequence $SSSS_0$ (that ultimately is equivalent to the 21-time steps of the sensory stream). In a sequential approach it took 21-time steps (assuming no latency) of the first level to reach the final identification of the sequence $SSSS_0$. By cascading predictions using this "trans-thalamic pathway" (if *L6+* and *L5* know the sequence) we can achieve sequence identification in the 3$^{rd}$ rung eleven-time steps ahead. After an *L5* prediction is forwarded, *L23* and *L1* can be disabled (since sequence identification is provided by the *L5* output). To do this, we simply mute *L4* input. This fact is represented in the figure as transparent labels in the *L1* output.

A subtle problem is that even if the replicas in *L5* and *L6b+* match, the size of the common bits may be much larger than the speculated symbol (i.e., we are predicting the union of multiple symbols). To prevent this, forward prediction is only done if the size of the resulting match is between certain limits (i.e., the match is narrow enough to represent a very limited number of symbols). In the most common case, if the size of the match is larger or smaller than the output SDR of the CC, the prediction is discarded.

When *L5* predictions are used recurrently, the process stops as soon there is a disagreement between *L5* and *L6b+*. In this case, the predictions of $SS_1 \rightarrow SS_2 \rightarrow SS_3$ in *L5,1$^{st}$* rung stops predicting ahead in the last value because is an end-of-sequence. Most likely *L6,2$^{nd}$* rung will have no prediction available to match with in the CTLoop. Then, the 1$^{st}$ rung is back in normal mode (i.e., L4 input is unmuted). This output is considered in *L6a* of the 2$^{nd}$ rung until EOS reaches the 3$^{rd}$ rung CC. During the forward predictions, the hierarchy works in decoupled stages. While the upper rungs advance in time using predictions, the lower rungs work by having a "ready to go" sequence identification when the dynamics of the higher rungs are disrupted. In a well-known input stream this behavior may occur pervasively thought the cortex (with constant activations and deactivations of supra-granular layers). An unknown sequence at a certain rung will be detected because its *L6b* will exit from the *known* state. In this case the supragranular layers are reactivated (i.e., *L4* inhibited). To prevent learning partial sequences thought the transitory, *L6a/b* learning will be disabled during *L4* inhibition. This is consistent with the fact that *L6a/b* learning is disabled when *L23* is in the *unknown* state, which will be the case when *L4* is inhibited (See Section 7). Although at lower rungs of the hierarchy the disambiguation capabilities matching *L6a* and *L5* predictions will be low due to the multiplicity of trajectories (i.e., wide predictions), at higher rungs this mechanism will be more useful. This scheme may be consistent with the translaminar inhibition networks originating in L6 corticothalamic cells [71].

This idea is inspired in the high speed of some high order recognition tasks [72] or the capability to tolerate sensory variation, via perceptual restoration [73]. The duality in the activation of *L5* and L23 when the sequence is predicted or not has been already described in [74]. The inhibitory effects of L6 on the cortical column [75] can be related with this. In these cases, cortico-thalamic loop seems to condition the activation of interlaminar parvalbumin cells (IL-PV) in L6a, which have inhibitory effects in L23/L4 and L5a. Therefore, a thalamic match might activate IL-PV cells (note that the coactivation of local L6 pyramidal cells, which is required for thalamic match, will also influence IL-PV), which inhibits superficial layers in the cortical column, as our model requires.

The learning coupling between *L6a+* and *L5* is loosely inspired in [76] observations in infragranular feedback connections between V2M/L and V1.

In any case, given the event rate in each layer, a higher level of activity is expected in the superficial layers than in the deep layers. This might be consistent with the layer-specific rhythm observed pervasively through the cortex; from sensory cortices [77] to PFC [78]. Although at the bottom rungs of cortical hierarchy these observations might fit into Predictive Processing Theory (PPT) [79], at the top rungs, in contrast to our model, they seem less plausible.

We speculatively hypothesize that, in biological systems, the proposed model corresponds to one of the functional purposes of the trans-thalamic pathway [14]. The high order nuclei of the thalamus receive L5 output, and the corresponding thalamic relay nuclei (TRN) receive L6a+ output. Only when the two matches do relay cells send the L5 activations to the corresponding cortical column, otherwise they are inhibited. Relay cells are excited only if there is sufficient input from L5. We hypothesize that all replica output from L5 will be aligned in the corresponding relay cell. In our case we force all replicas in L5 to agree before activating the relay cell, but in a biological system a certain threshold will be most likely used (and most certainly many more L5 replicas). Note that despite being far apart from the target the thalamic relay cells have the fastest axons in the brain [80], which is consistent with the functionality of the proposed model. Having multiple replicas L6a+ might increase the accuracy of this filtering. Note that L6a$\rightarrow$TRN synapses are metabotropic, which requires a steady activity to disinhibit the relay cell [56].

The same technique could be used by first order thalamic nuclei, such as de LGN or MGN to prevent noise coming from the brainstem. Although presumably certain specialization should be carried out by brainstem nuclei, CC in A1 or V1 will condition MGN and LGN respectively reducing the chances of sending unexpected activations from visual and auditory pathways. We will address this, for the case of MGN, in Section VI.B.

*L5* output may be useful in a variety of subcortical structures not only to control motor output but to enrich and/or stabilize sensory input. Collectively, L5 predictions from all cortical columns represent a snapshot of the state of the cortex. Our speculative hypothesis is that L5 efferent copies from motor and non-motor cortices are used by brainstem nuclei to modulate both input and output from the cortex.

As per the model, even if there is a match between *L5* and *L6a+*, the actual symbol determined by *L23* in the absence of CTLoop may differ (i.e., the speculation is incorrect). From a system behavior perspective, this discrepancy is inconsequential as both cortical columns (CCs) will consistently utilize the presumed "incorrect" values. The sole requirement is the continuous activation of CTLoop; a deactivated CTLoop would impede information recall. This aligns with the observed memory deficits resulting from damage to specific higher-order thalamic nuclei [81].

The proposed loop can be seen as a bottom-up attention mechanism, in the sense that it focuses on pattern identification, locally via the thalamus (L6) and globally via brainstem nuclei (L5) mediation. Finally, it should be noted that this information may be useful not only to enhance input and output, but also for driving Basal ganglia circuit loops. Some L5 subpopulations send efferents to the striatum [82]. Section V.B propose a with top-down attention mechanism using this idea. Moreover, Section V.D foresees a potential use of the L5 expressed cortical state to perform supervised learning.

Although recurrency in L5 can be consistent with the fact that in biological systems recurrent connections within L5 are more frequent after the training [83], we could not find any clear pathway between the thalamus and L5 to filtering out in fraction of the recurrence that do not match with L6a+ expectations. Unfortunately untangling the complexity of cortico-thalamic-cortical loops is a work in progress for current neuroscience [84].

In any case the CTC interactions proposed by the model support the idea that higher rungs of the hierarchy can operate loosely coupled to the senses, as has recently been demonstrated by [85] for the case of speech processing. The parallel processing between primary auditory cortex and speech processing fits well into our model, explaining how it might work and why the former needs to be bootstrapped from the latter. Therefore, our model reconciles the puzzling contradiction between the experimental results of [85] and the hierarchical organization of the cortex. In complex systems where regions may be reciprocally connected, L5 input recurrences can give rise to autonomous (i.e., non-input dependent) activity. This may be related to reentrant signals observed in biological systems [86]. Therefore, the discussion goes far beyond the complexity of the problem at hand, as such activity could be the supporting phenomenon of much more ambitious aims [87][88]. From a distance, our model is compatible with how shared thalamic bridge allow phenomena such as [89].

IV. HIPPOCAMPAL INTERVENTION

A. *Filtering Spurious Data across Transitory Learning*

When a group of CCs in a rung is exposed to a novel input, sequence learning may occur out-of-order. Therefore, the CCs in the next rung will be momentarily fed with spurious data. This noise can propagate through the hierarchy, learning many non-existent sequences. These sequences increase the system's synaptic load, the time required to reach a stable state, and decrease its accuracy (i.e., increase the width of the predictions).

The solution to this problem is to use a mechanism to detect when the current rung has a "good enough" model of the input, and only then allow the CCs in the next rung to learn the input. By introducing an ephemeral *Sequence Memory* between the rungs (i.e., inserting it on demand and routing the output of the current rung accordingly), it is possible to address the issue. This component is referred to as **_Filtering Slice_** or *CA3*. The exponential moving average (EMA) of the anomaly score for this component, using a small alpha value, provides an estimate of the stability of the output produced by the CC in the rung. Hence, if *CA3* is in *Unknown* state we turn off the learning of the affected CCs. While the transient nature and routing capabilities facilitate the reuse of the same *CA3* through multiple cortical columns, there needs to be a mechanism in place to initiate the allocation process from a common pool. We will utilize the learning state of *L6a* in the CC of the next rung. If one of these CCs enters the *Unknown* state, *CA3* will be allocated, and consequently the learning processes of the upstream CC will be disabled.

Figure 4, shows the implementation of this concept in a 2x2 cortex. Each rung has two CCs and there is a cross-effect in the symbol values produced by each one. This is known as lateral contextualization and will be discussed in detail in Section V.A. It is possible to combine the outputs of $C_{00}$ and $C_{01}$ over a single SM or multiple SMs. As with any layer of the CC, it is possible to replicate the SMs to increase prediction accuracy if needed. When multiple SMs are in use, the state of CA3 is determined by the result of the vote of all individual SMs (e.g., half plus one).

CA3 allocation occurs when either *L6a* in $C_{10}$ or $C_{11}$ moves to *Unknown* state. The presence of *CA3* does not disrupt the flow of feedback between rungs. *CA3* effects vanish when both *L6a* return to the *Known* state. This concept can then be understood as a local mechanism for detecting new data and selectively activating learning when it is safe to do so.

This component can also be useful in preventing noise-induced changes in the input segmentation layer while the cortical columns are exposed to known data. The learning cycle of *L6b* responds rapidly to input changes, and persistent noise may cause alterations. To avoid such changes in the synapses, learning on *L6b* is only enabled either if the *CA3* at the input or at the output of the CC is allocated.

Furthermore, output instability of $CC_{0x}$ output also affects *L5* in $CC_{0x}$. To address this issue, *L5* used the same rules used in the *L6a* of $CC_{1x}$, which means that learning will only occur if *CA3* and *L23* in $CC_{1x}$ are in the *Known* state.

The learning *L1* and *L4* is also gated by the presence of CA3 above or below. *L1* can learn only if the CA3 below is present and in *known* state and, learning stops once CA3 below vanishes. If a CC does not have a CA3 above or below, learning is disabled in *L1* and *L4*. This allows to freeze the projections once the CC is stable. Figure 5 summarizes the rules of learning gating.

### B. *Replay as a Mechanism for Accelerating Cortical Column Learning*

The learning hysteresis cycle of *CA3* mantains the synaptic load constrained, even though *CA3* lacks a sequence segmentation mechanism like SM in conventional CCs. The synaptic content of *CA3* can be gradually pruned after deallocation. Therefore, assuming a shared structure there is a certain urge to deallocate *CA3* as quickly as possible. Unfortunately, due to the dimensional reduction performed in each rung, it may take a long (physical) time for all *L6b* above to reach the *known* state. Conveniently, once *CA3* is in the *known* state we can repeatedly use its predictions to speed up the process.

By driving *CA3* with its own predictions (i.e., using recurrent inputs, as we did in *L5* to predict multiple symbols in the sequence), we can quickly recall the remaining values of the sequence. By sending these predictions to the appropriate CCs, we accelerate learning because the delay between inputs will be minimal. This fast sequence will resemble the hippocampal *Short Wave Ripple (SWR)* [90] [91]. Multiple replicas in *CA3* can be used to increase the accuracy of the recurrent predictions. Thus, a bit in the SWR is only used if the number of replicas above a certain threshold agrees whith its prediction.

In most cases, the SWR is short, usually matching the sequence length according to the partitioning that the next rung will discover later. Although this data may be inaccurate, and potentially conflicting with the real

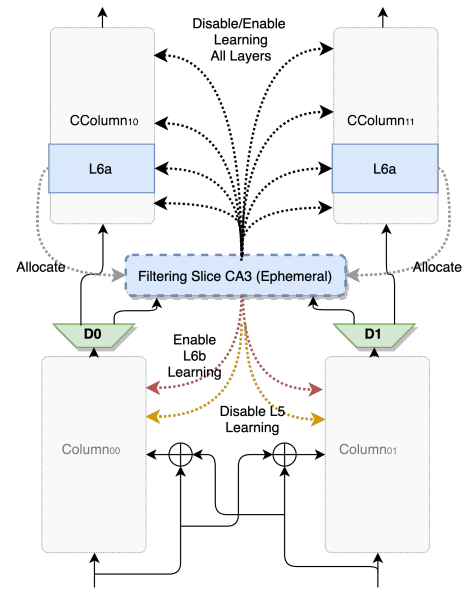

Figure 4 CC Filtering. 2x2 Cortex Example

```
if CA3 is present then:
  if CA3 in unknown then:
    Learning in CC above   disabled
    Learning in L5 below   disabled
  else:
    Learning in L1    above  enabled
    Learning in L23   above  enabled
    Learning in L4    above  enabled
    if L23 above in unknown then:
      Learning in L6a/b above  disabled
      Learning in L5    below  disabled
    else:
      Learning in L6a/b above  enabled
      Learning in L5    below  enabled
    endif
  endif
else:
  Learning in L4/L1/L6b    above        disabled
  Learning in L23/L6a above & L5 below  enabled
endif
```

Figure 5 Learning Gating Algorithm

data, since the SWR occurs simultaneously, the next rung is partially exposed to the sequence for a short period of time. Therefore, to maintain the stability of the *L6* sequence segmentation, we should avoid mixing real and replay data. Real data is sent to *L6* and *L4*, and replay is sent to *L4* only. This design choice seems to be consistent with the fact that hippocampal afferents in the cortex are mainly located in the supragranular layers [92]. Therefore, SWR can be understood as a mechanism to accelerate CC learning by speeding up *L4* clustering convergence. To prevent sequences with mixed real and replay data from being learned by *L23*, the *L4* output is not sent to *L23*. Moreover, the biological plausibility of sending the replay to *L23* is reduced because the time constants in all the layers of the CC are fixed.

To prevent excessive *L4* clustering if the *CA3* prediction is wider than the union of two symbols (e.g., if the unknown threshold in *L6b* is 0.5, then the maximum number of bits tolerated in the prediction is 150% of a regular symbol), L4 input is inhibited. As soon as a prediction is not wide enough (i.e., fewer bits than the sparsity level) or CA3 receives a regular input, the SWR is interrupted. If the predicted value of CA3 is equal to its input, the SWR is interrupted. To prevent data contamination in *CA3*, learning is disabled during SWR.

Although the effort required may seem considerable for an apparently small benefit (i.e., faster *L4* learning), empirical evidence suggests that the effect is significant. Slow convergence of *L4* clustering significantly increases the synaptic load on *L23* and the time required to stabilize the hierarchy upwards. The gains will be more noticeable when the L4 learning rate is small.

### C. Long-Term Learning Modulation

*L6a* can enhance the learning ability of CCs beyond their role as drivers of *CA3* intervention. The level of knowledge of *L6a* will modify the *K* parameter depicted in Figure 2, which affects the sharpness of the hysteresis cycle, as well as the LTD/LTP increments in the synaptic weight of *L23*, *L5* and *L6b*. The parameters will be multiplied by a modulation factor that depends on $AEMAAS_{L6a}$. The idea is that if *L6a* is in the *Unknown* state, we have a stream of new data that should be learned as fast as possible.

Indirectly, the state of *L6a* determines the presence of a *CA3* below, and *L6a* on the next rung (*L6a+*) determines the presence of a *CA3* above it (*CA3+*). Thus, *L6a* will modulate the learning of *L23* and *L6b*, and *L6a+* will modulate the learning of *L5*. Then, for a given modulation *m,* the learning parameters change as (where d is the increment of permanence for LTP and K is the steepness constant in logistic functions of the hysteresis cycle):

$$K= (1.0 - m)·1000$$
$$d=d_0 · (1 + m·100)$$

We will also use a hysteresis cycle of two logistic functions for *m.* Then,

$$m = \begin{cases} \frac{1}{1+e^{-1000(AEMAASL6a-Down)}} \text{ in } Unknown \text{ state} \\ \frac{1}{1+e^{-1000(AEMAASL6a-Up)}} \text{ in } Known \text{ state} \end{cases}$$

Although this is optional, it improves learning speed significantly at the expense of a slight increment in the synaptic load.

### D. Using CA3 State to Guide Active Pruning

An open problem that *CA3* can also help to solve is the active pruning of *L4* and *L1* dendritic segments. To increase system stability by minimizing the risk that new incoming data will alter the current representations (i.e., to reduce the chances of interference between the already known model and new stimuli), the key is to remove the unused initial silent synapses in *L1* and *L4* (i.e., to destroy output dendritic segments in *L4* and *L1* that are not actively being used).

The problems we face are twofold: when to do active pruning and how to do it. Considering the *CA3* functionality *when* is straightforward: when CA3+ are deallocated it is safe to perform it. This means that the CCs in the next rung know the sequence, and then we can consider the output of the current CC to be stable. This is intuitively equivalent to starting to *freeze* the output symbol diversity.

Although the *how* can be complex, we opt for a biologically plausible solution. If a proximal segment has been sparsely used (or not used at all), it will have unconnected initial synapses. Proximal synapses cannot be created throughout the lifetime of the system. Therefore, both *L4* and *L1* must have a topologically rich connectivity at boot time, with a significant proportion of silent synapses (e.g., 70% of existing connections, with 80% of them being silent)[5]. In each actively used segment, heterosynaptic plasticity will implicitly remove the silent synapses. The process is then as simple as progressively eliminating all segments with at least one silent synapse. The removal rate is a design parameter of the system. From a biological perspective, this may be plausible since activity mediated "don't eat me" signals, such as CD47 [93], prevent microglia from active phagocytosis. Our empirical evidence suggests that active pruning is necessary to achieve a stable system. Apparently, by eliminating silent synapses, we reduce the learning interference of new sensory data with the already known model. This is consistent with the "goal-directed" convoluted mechanisms that evolution has sculpted into biological systems [94]. However, it contradicts the commonly attributed functional role of silent synapses in the formation of new memories [95]. According to our model, they have no functional purpose. They are just a remnant of learning. Recent experimental evidence, such as [96], may support this view.

### E. Biological Inspiration/Plausibility

We speculatively hypothesize that CA3 performs the previously proposed tasks. The main functional role of the remaining parts of the hippocampal complex (i.e.,

---

[5] This means that there is a synapse between any output and 70% of the inputs, with a 20% probability of being connected (e.g., permanence above threshold).

CA1, DG, EC, etc..) and possibly other specialized cells through the cortex, is to associate the region of the cortex that CA3 support for learning. This circuit looks like a cache in the memory hierarchy of a conventional computing system, where each set can store multiple real addresses from the main memory. Each cache copy of the main memory has a TAG associated that is used to disambiguate the real address when the processor needs to access it. We speculatively hypothesize that CA1 represents the TAG array (i.e., which CCs in the cortex are associated) and CA3 represent the DATA array (i.e., where to track the input data of such CCs). The remaining elements of the hippocampal formation, i.e., subiculum, dentate gyrus, EC, etc. are dedicated to routing the information to/from the cortex and CA3. This may explain the co-fluctuating activity in the hippocampus and certain cortical cells [97] and the observed dual-plasticity properties [98]. Contrary to the currently most accepted model, we speculatively hypothesize that grid cells, place cell, etc… are only used to route the information to/from the cortex and hippocampus. Hence, experimental evidence about the grid/place/border/head/object… cells in the hippocampal complex (or other cortical regions [99]) may be epiphenomenal. According to our hypothesis any cortical column should be able to use the hippocampal complex, requiring the existence of "addressing" cells through cortex, EC and CA1 to route the input and output to and from CA3. The spatial representation of such cells is not related to the physical environment of the biological system or to any cognitive construct, but only to the location of the cortex where the filtering, modulation and replay takes place. This idea is partly inspired by the fact that the variation of grid cells in the EC when moving from ventral to dorsal regions of the Medial Entorhinal Cortex (MEC) [100] is strikingly similarity to the variation of the index bit in a memory address bus (from less to most significant bits). Grid cells activity is also stable from the beginning of the training (even when the environment is novel [101]) and is then independent of the cortical input (i.e., the data). Therefore, dorsally located cells might encode the position of large regions of the cortex where the cortical column might be encoded and as they move to ventral regions, the cells encode potential sub-regions with higher precision. When all are combined, the actual cortical column, as occurs in an address bus, is completely disambiguated. Similarly, the Lateral Entorhinal Cortex (LEC) appears to act as a data bus [102][103]. The information sent from the cortex to CA1 could represent the "where stream (ventral)" and to CA3 the "when stream (dorsal)". This model could explain the two stream hypothesis [104]. According to our model the main purpose of the hippocampus is processing temporal information from the senses. In other words, there is no spatial information in the sensory stream, just cortical "addresses". Although controversial, this idea has already been proposed [105].

In the SWR process, inspired by the biological evidence [106], we choose to feed the entire prediction as recurrent input. This is inspired by the rich and complex effects that the replay from the hippocampus to the cortex has on biological systems [107]. The evidence for backward replay [13] seems inconsistent with the proposed model. Our speculative hypothesis is that the hippocamp algorithmic construct serves both for cortical activity originating from sensory perception (as assumed in this paper) and corticostriatal loops. Some of these loops, thought to be responsible for the reward system [108], may simulate sequences in the cortex to improve the biological system's learning outcome. For example, learning from predictions to reach the behavioral goals faster. Assuming a fixed learning algorithm for real and simulated information, such replays should be treated as sensory information. Therefore, SWR should be activated to accelerate spatial pooling in L4. Thus, the experimental correlation between the reward system and hippocampal replay [109] may be epiphenomenal. Learning gating and long-term modulation guided by CA3 activation patterns is loosely inspired by the cholinergic system. Besides direct synaptic plasticity effects (neuron and layer specific [110]), acetylcholine modulates memory formation through glia regulation [60]. Our speculative hypothesis is that the effects of the cholinergic system are: (1) it increases synaptic efficacy by improving the excitability of pyramidal neurons [111], (2) it alters the chemical dynamics of astrocytes, increasing the release of certain neurotransmitters in the PAP [112], and (3) it also regulates the calcium concentration in the synaptic cleft. Finally, the specificity of L5, appears to be weakly consistent with the cross-layer learning modulation heterogeneity observed in biological systems [113].

This seems to complement the "short-term" modulation hypothesis introduced in section III.A. Perhaps not coincidentally the cholinergic system appears to have both fast and a slow effects on the cortex [110]. From the perspective of the hippocampal complex and the intricate connectivity of the septohippocampal pathway of the fornix [114] this mechanism could be fundamental in "activating" the learning in the corresponding region of CA3, where the data routed from the cortex must be temporally learned. Hence the coactivation of the system in certain regions of the cortex and hippocampal complex may act as a scaffold to enhance the learning rate of new data, while enhancing cortical stability when the data is known. Although the anatomical complexity of the Basal

Forebrain (where most cholinergic neurons are located) prevents the establishment of clear input-output relations, [115] seems to indicate that there are hippocampal effects through the interposition of the lateral septal nucleus. In any case the complexity of the cholinergic system seems to have overlapping functionalities.

Our interpretation appears to fall between the Standard Model of Consolidation (SMC) [116] and Hippocampal Indexing Theory (HIT) [117]. While SMC fails to account for retrograde amnesia, our model provides an explanation. Without the proposed hippocampal mechanisms, integrating previously acquired knowledge with new information will fail (because learning gating may be unavailable). Without this feature destructive interference between known and new data is likely. Instead of integrating new facts in previous knowledge, this might destroy it. If the damage in the hippocampal complex is partial, the destructive interference will occur in the region of the cortex where the functionality is not available (i.e., there are no routing possibilities), so it can be selective. Therefore, and contrary to HIT (and other memory theories), the hippocampus does not contain any long-term indexing information about the cortical content, it only contains the cortical topological mapping necessary for controlling long-term learning modulation, learning gating, filtering, and replay.

The set of speculations has a few pieces of weak plausibility. (1) The age-related heterogenous activity in the traverse axis of the CA3 [118], with proximal activity being more prominent in early age (presumably used when the lowest rungs of the cortex are learning) and distal activity being more prominent in adulthood (presumably used when the learning is performed in the upper rungs of the hierarchy). (2) According to research, the posterior hippocampus is more closely linked to detailed facts, while the anterior hippocampus is more closely linked to higher order constructs [119]. This also suggest some correspondence with the cortical hierarchical organization (i.e., there is a limited associativity, in computer architecture terms, between the cortex and CA3). (3) The proximal region has fewer recurrent axons [120], suggesting that SWR may be less attractive in these areas due to short inter-arrival times. (3) In contrast to CA1, CA3 is characterized by *oriens lacunosum-moleculare* (O-LM) interneurons with large longitudinal axonal spans that support the presence of cell assemblies [121]. This appears to align with the individual Sequence Memories per input stream required by our model.

Our hypothesis regarding the mapping between the cortex and hippocampus is like the concept of limited associativity in a processor cache. Full associativity, where a memory address can be stored in any cache position, is too expensive. Therefore, each memory address has a limited number of positions in the cache where it can be stored.

V. THE ARCHITECTURE OF THE CORTEX

A. *Lateral Contextualization*

Once the cortical column and auxiliary structures are laid out, it is important to consider the input mapping and how the internal topology of the cortex is organized. For the inputs, given the fact that sensory cortices match the topological distribution of the sensor [2] (e.g., retinotopic in V1 or tonotopic in A1), each input stream[6] is used independently to build the temporal reference frames of each CC in the cortex's front-end.

Sensory cortices are characterized by a lower number of pyramidal cells (ergo fewer synapses) [2]. Our hypothesis is that multiple input streams are "combined" only upward in the hierarchy and therefore sensory cortices require fewer synapses. Since the neuroscientific knowledge of the upper rungs is less clear, we should fall back on the principles of logic. The key argument for the definition of cortical connectivity is the improvement of identification abilities through the enrichment of symbol expressiveness in each CC. Additionally, the nature of the input to the CC may be heterogeneous (e.g., manifested as different sequence lengths in different tones or topographic zones), so we need a mechanism to regularize the activity across CCs. To achieve this, the sensor's information can be progressively distributed to more CCs in successive cortical rungs in a self-timed manner. Then the influence of each input spreads laterally across the rungs of the hierarchy, further away from the sensory cortices [14]. The mechanism should use a limited number of long connections to be consistent with the biological distribution of axonal lengths [122]

Figure 6 illustrates the concept for a 1D cortex. Each CC is divided into *N+1* independent modules. The input ports of each module (i.e., *L4* and *L6a/b* layers) are fed by distinct CCs in the preceding rung. The L6a input corresponds to the aligned CC (i.e., $C_{10}$ in this case). In contrast L4 input comes from the aligned and adjacent CC (i.e., $C_{10}$, $C_{00}$, $C_{02}$). Intuitively, the output symbol's temporal reference frame is constructed using the preceding aligned CC's output. The output symbol's value is determined by the combination of the sequences generated by the preceding aligned adjacent and CCs during the temporal reference frame. The

---

[6] An input stream refers to a portion of the sense, such as frequency-dependent activity in auditory cortex or activity from a region of the retina in the visual cortex.

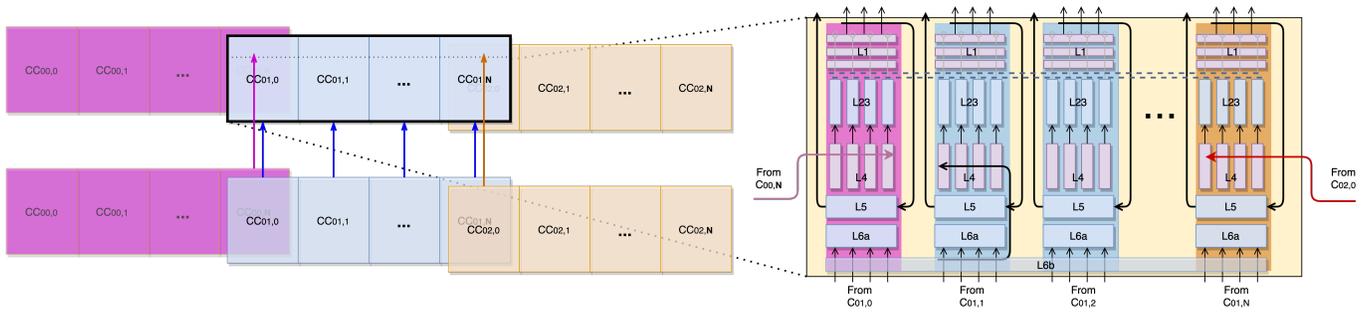

**Figure 6 (left) Cross-coupling of the cortical columns in a 1-D cortex, (right) Sub-module and input/output selectivity in the cortical column.**

rationale for not feeding *L6* in an analogous way is that the timing of the CCs involved is not aligned (i.e., the EOS in $C_{10}$, $C_{00}$, $C_{02}$ usually does not happen simultaneously). We denoted this mechanism as *Lateral Contextualization*. In this example, the modules work as follows:

(1) Each inner-replica (four in the example) of the input module of the CC affects an independent *L4/L23*
(2) Each module (N+1 in this example) has at least one completely independent *L6a*. The knowledge status of the CC (and long-term modulation) is established by the majority of the status of all the replicas. Short-term modulation is done at SM level.
(3) To improve the stability of the EOS, multiple modules can share a single *L6b* layer (all in the figure). The short time constant of *L6b* (i.e., alpha close to 1) moderates the synaptic load of the layer, despite the requirement of a larger receptive field.
(4) In EOS, the temporal pooling is performed by each module's *L1* replica, which samples the predicted state of connected cells in *L23* of any module. The connectivity of active and silent synapses from *L23* replicas to each *L1* is set at initialization.
(5) Each module's output value is fed into its respective *L5*. If attention is present, subsequently *L5* prediction is sent to CTLoop logic. To increase the accuracy of the matching procedure, multiple replicas of *L5* may be present per module.
(6) The CC may optionally use only a single lateral contextualization.

Instead of using one module per incoming flow, we can increase the resilience and capability by using two or more. The configuration of the CC (number of modules, incoming flows, internal replicas, etc..) can be dependent upon the rung of the hierarchy where the CC is located, the nature of the input or the reliability of the substrate. In our proof of concept, a homogenous cortex is used with no internal replicas in the modules and a single input flow.

We speculate that such modules might correspond to the micro cortical columns observed in biological systems [2][123]. From a developmental perspective, this sub-hierarchical organization of the cortical column may be less demanding to achieve cortical region specialization. Lateral contextualization does not allow the use of large receptive fields in *L23* because the input may not be synchronous. Since each *L23* projection receives from potentially different, presumably out-of-sync, CCs, each input stream must be tracked individually. In fact, the contextualization is inspired by well-known facts about the heterogeneity of *L23* and *L6* receptive fields [29].

After defining the CC modular structure, the cortex structure can be laid out. Figure7.A shows the forward connections in a linear (or tonotopic) cortex, which has four rungs in the hierarchy, and four input streams involved (i.e., a 4x4 cortex). Each encoder is fed with a set of features from the sensor, presumably using a localist representation, and transforms it into its SDR representation. Section VI.B will discusses how such encoding is performed in the case of an audio signal in the speech recognition problem.

Cortical connectivity should enable each input to have effects in an increasingly larger number of cortical columns as one moves up the hierarchy in a scalable manner. First, the symbol generated by one CC will incorporate sequences from several CC in the previous rung. Therefore, a CA3 slice will be inserted in each rung to ensure that all CCs below are providing stable information. In a large cortex, a unified CA3 slice per rung is not biologically plausible, as CA3 requires knowledge synchronization across the sequence memories that comprise it. *B* cortical modules will be used in forward connections and *b* in laterals. To maintain scalability of the CA3 architecture, lateral contextualization is limited to a single CC. This constraint necessitates one CA3 slice for every two CC groups. To gradually distribute the impact of each input, a slice and connectivity shift is implemented. Wraparounds are utilized to maintain system symmetry.

Figure7.B shows the influence of each one of the four inputs on each cortical column of the cortex, for both *L4* and *L6*, using color coding. In the example, *L6* of $CC_{12}$ is only affected by *Input$_2$*, while *L4* is affected by *Input$_3$* and *Input$_2$*, and therefore also affects *L6* in $CC_{22}$, and so on. The degree of influence of laterals in the *i* rung is determined by the ration between the forward and lateral widths. In practical systems, it is reasonable to use a low rate (or even zero) in the lowest rungs and gradually increase it as we move up in the hierarchy. This approach ensures that the synaptic load at the bottom of the hierarchy is not excessive, and the disambiguation capabilities are improved at the top.

Although the overlapping selectivity shown in the example is limited, we can go up to 3-to-1. A bidimensional cortex, assuming a hexagonal shape for the CCs [2], will have the opportunity to have 7-to-1 (see Figure 9). Then, the extension of effects across the cortex of primary tonotopic cortices will require fewer hops.

It is important to note that in *L6*, there will be no lateral contextualization (i.e., $CC_{ij}$ will only be fed with $C_{(i-1)j}$ *L1* replicas). However, *L23* determines when it can learn (see Section III.B). Consequently, even if sequences of neighboring CCs are not temporally aligned, *L6* will learn within the underlying periodicity of its context. According to this, *L6* constructs a temporal reference frame for the CC at the point where identification occurs, which is weakly influenced by the lateral horizontal context.

Although contextualization across rungs (i.e., vertical contextualization) is not considered in the design, it could be used. Then it could be possible to feed certain *L4* modules with outputs from CC two or more rungs below.

Our speculative hypothesis is that biological systems use similar constructs, not only locally but also between distant regions. In fact, considering certain L23 and/or L5a pyramidal cells, axons project into the intratelencephalic tract (IT), affecting distant cortical regions [124] [125]. Potentially L5a is functionally equivalent to L23 but with higher chances of affecting contralateral regions. This can be used to assemble a modular cortex using specific connectivity through preselected groups of cortical columns (i.e., non-regular connectivity). However, given the simplicity of the input considered in this work, we did not require the definition of cortical regions. Consequently, we only consider regular connectivity through columns and do not address the hypothesized role of L5a. A multisensory cortex may necessitate such a heterogeneous architecture. Furthermore, we speculatively hypothesize that the concept of lateral contextualization is crucial in addressing the *Temporal Multisensorial Binding Problem* [126] in biological systems.

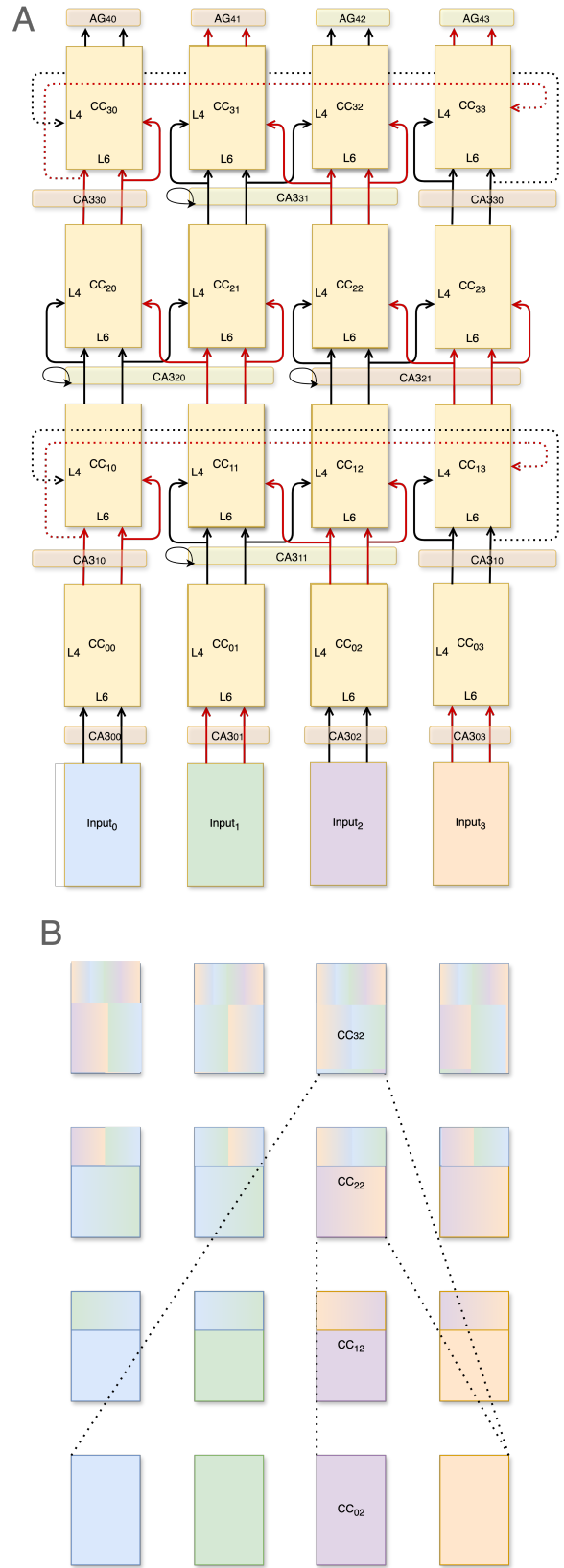

Figure 7 A: Forward Connections of a 4x4 Cortex.
B: Input Contribution to the L6 and L4 Layers of the Cortical Columns

Moving up in the hierarchy, disambiguation and resilience may require increasing the identification capabilities of each CC. Presumably, patterns will be more diverse and/or failures could have a greater impact on system functionality. This may suggest that it is necessary to increase the complexity of the cortical column, for example, by increasing the synaptic load through cellular specialization. This solution may incur a significant genetic cost, which could be observed in cross-species gene comparisons. Our hypothesis is that the system could maintain these costs using a **modular** approach. In Figure 6, it is possible to identify multiple largely independent modules. Increasing the number of modules per CC increases capacity. Regardless of its position in the hierarchy, each module should have a similar synaptic load. We speculate that these modules may correspond to the microcortical columns observed in biological systems [2].

According to the lateral connectivity, it is necessary for multiple CCs to share the same CA3 slice. In this way, noise generation due to the different speed in sequence learning in the source CCs is avoided. For instance, if $C_{00}$ acquires new sensory data before $C_{01}$, we delay learning the corresponding sequence in $L23$ of $C_{10}$ until a reference frame is available (i.e., $L6$ input is available from $C_{00}$). The absence of EOS will substantially increase the synaptic load of the affected group of $L23$ in $C_{10}$.

In areas of the cortex of biological systems where multiple sensory inputs converge, coordinated filtering can be extensive. We speculatively hypothesize that specialized structures, such as the claustrum [127], facilitate this synchronization. The claustrum has extensive connectivity with sensory cortices (from L6) [128] and the hippocampal complex [129], which supports this conjecture. This, along with hierarchical organization, may reconcile the apparent discrepancy between interstimulus interval effects in behaviorally measured associations and neurological LTP evidence [130]. In cortical regions where direct link with behavior can be observed, certain events might elicit the prediction of stimuli that are not directly related. If the flow of information converges on a common region, the interleaving of events will condition the behavioral output of the model. Furthermore, the relationships established in converging regions, may provide an explanation for the perceptual binding problem [131].

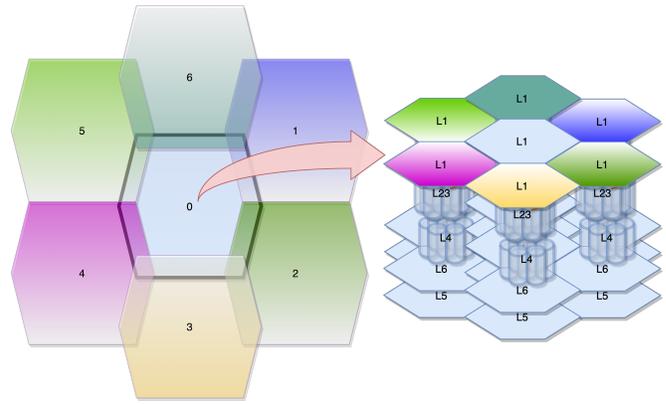

**Figure 9 Cross-coupling Cortical Column in a 2D Cortex**

### B. Attention

The CTLoop algorithm interacts subtly with lateral contextualization. If only one CC in a rung receives L5 recurrent predictions from the previous CC, it will progress faster than the neighbors. Although the *Sequence Memories* use only relative order to predict values (i.e., ignore precise timing), at any given time, lateral inputs of L23 will be out of sync with L6 (especially if CCn are not using CTLoop). This will modify the symbol representation, as lateral inputs may shift to the next or the previous sequence. To prevent this from happening, it appears that CTLoop must be performed synchronously. At first glance, addressing this issue may not only necessitate a biologically implausible mechanism but also significantly decrease the likelihood of having predictions. Additionally, producing forward predictions only in the lower rungs may not be beneficial if the upper rungs cannot accelerate the output generation. In this scenario, the forwarded predictions are wasted because pattern identification cannot be expedited at the top of the hierarchy. Therefore, it is reasonable to disable the CTLoop (e.g., prevent the L5 output from leaving the cortical column) if some of the affected cortical columns in the next rung (i.e., those where the current output is used as L4 input) are not forwarding predictions. If we do so, there is no need for simultaneous prediction in neighboring CCs. If all the affected cortical columns are forward-predicting, since L4s are inhibited, we can combine predictions with real data without altering the output representation, which is being generated by predictions.

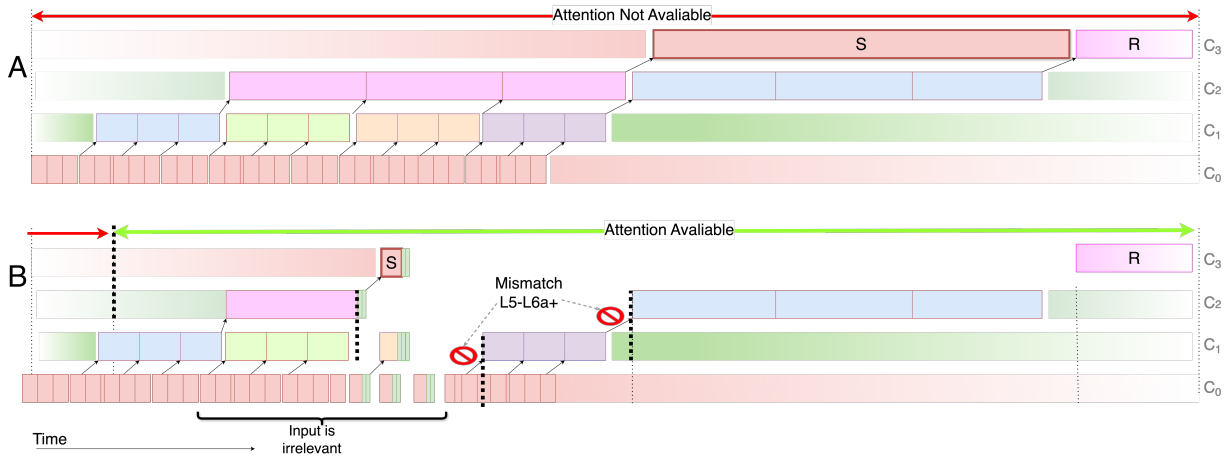

**Figure 8 Enabling of CTLoop A: Attention is not available, i.e., (AG is now aware of the context because internal CA3 is allocated), B: Attention is available (internal CA3 of the AG is deallocated).**

We introduce the concept of *Attention*, which models the fact that top-level CCs of the cortex are in the *known* state. To implement this, we introduce an additional CA3 slice per CC at the top of the cortex, along with the logic of allocation, denoted as Attention Generator (AG). For example, in Figure 7, $AG3_{4x}$ are the "attention generators" for the preceding CCs. The intuition is that we only have attention if the output stream of the corresponding top-rung CC is fully aware of the input, which is determined by the learning gating of its own L6b, hence we need an output CA3 and its allocation logic. Attention can be generated externally, but such a scenario is not considered.

Let us assume a 1x4 cortex at a time when attention is not available (Figure 8.A). In such a scenario, none of the cortical columns (from $C_0$ to $C_3$) can produce forward predictions. In contrast, when the CA3 in the AG is deallocated (Figure 8.B), the L5 output of $C_2$ is compared with L6+ in $C_3$. As discussed in Section III.D, the identification of top-level symbol *S* at $C_3$ will occur much sooner. Under these circumstances, downstream cortical columns $C_0$, $C_1$, and $C_2$ are gradually enabled to make forward predictions too. The prediction burst is interrupted when there is a mismatch between L5 and L6a+. Nevertheless, lower-rung mismatches may have no effect if the upper-rung forwarding is correct. For example, the $C_0$ to $C_1$ mismatches are disregarded because $C_1$ and $C_2$ previously agree on the correspond prediction. If upper-level mismatches happen, for example $C_1$ to $C_2$, the system could seamlessly reattach the top-rung to the real sensory stream, with no effect on the subsequent sequence *R*. When such situations occur, attention progressively disappears downstream until the top-level CA3 regains familiarity with the context. In addition to identifying patterns quickly, this mechanism also reduces sensory dependence. During forward prediction, sensory information can be noisy or even disappear. For example, in Figure 8, in the region labeled "Input is Irrelevant", no input will affect the upper CC. This occurs at every rung, which enhances the noise tolerance through the entire hierarchy. Thus, in addition to selective attention (and identification speed-up), this can be understood as an idea to increase the reliability of the cortex. Despite the simplicity of the example, it is evident that the primary advantage of the mechanism lies in the top rung. It should be noted that thus is where the most significant time reduction per prediction is achieved. In lower levels, the risk-benefit ratio appears to be low, which increases the appeal of the idea.

This idea is consistent with the fact that attention enhances synchronization across regions of the cortex (e.g., FEF and V4 [132]). Beyond this case, it is clear that the thalamus plays an important role in the attention mechanism [133][134], as our model requires for the CTLoop. Although our model of attention generation is simple, we speculatively hypothesize that biological systems use similar mechanisms most likely controlled by a high-level region of the cortex to focus attention on the most relevant part of the sensory stream. While only top-down attention is considered, we speculatively hypothesize that bottom-up attention is triggered when L5-L6a+ matches begin at the lower regions of the hierarchy (our example, assumes that they begin matching immediately when the attention is available). We can extend this discussion one step further. It is well known that attention (and awareness) increases the relevance of ECG oscillations throughout the cortex. We speculatively hypothesize that neural oscillations are an epiphenomenon of the described mechanism and have little to no functional relevance. The suppression of the input effects in upper rungs of the hierarchy may partially explain elusive cognitive phenomena such as the cocktail party effect (although it may be involved in many other perceptual effects).

Attention, in conjunction with the still unknown corticostriatal loop function, can also be used to control the output of the motor cortices by enabling only the corresponding L5 cells that are volitional to act on subcortical structures involved in action.

### C. *Feedback as a Mechanism to Stabilize Inhibition Processes*

As *L6a* predicts the next input, $CC_{(i+1)j}$ *L6a* predictions (*L6a+*) can be used as "guidance" for $CC_{ij}$. *L6a+* predictions will be consistent with the output of $CC_{ij}$ (i.e., with the *L1* inhibition results for the next input). By using this feedback to slightly adjust the overlap of predicted *L23* cells at $CC_{ij}$ EOS with each out bit towards the predicted values, we increase the stability of the output symbol. *L6a+* predictions can be used analogously in the k-WTA of the *L4* in $CC_{(i+1)j}$ and any other $CC_{(i+1)j\pm1}$ using lateral contextualization (i.e., with supragranular layers affected by the L1 output of $CC_{ij}$). In the presence of noise, this increases the likelihood of producing a k-WTA in L4 consistent with the prediction. Hence, feedback is considered as a modulatory input of *L1* and *L4.*

Figure 10 shows the feedback connectivity for the previous example (both L6+→L1 and L6+→L4+). Encoder considerations are discussed in Section VI.B. *The maximum overlap contribution* of all modulatory inputs is always less than the overlap that a single connected forward input can produce. Therefore, an output bit cannot win the k-WTA if there is no active input bit connected. According to this, the purpose of the feedback input is resolving ties in the overlap, which is relevant when multiple output bits with the same total number of active input bits are connected[7]. Intuitively, feedback can be understood as a modulatory mechanism to "stabilize" the internal state of the CC. While *L1* is learning, it opens the opportunity to steer symbol representation (i.e., enables some form of supervised learning). The ability to control the inhibition results depends on the size of the ties (and consequently the number of output bits with an overlap with the input large enough to be selected). The size of the ties is determined by the initial connectivity of *L4* and *L1*. Thus, if connectivity is relevant, and the seemingly minor effects of feedback are magnified. Our empirical results suggest that a minimum of ~20% connectivity from each input to any output is needed to be able to disambiguate complex patterns.

Since *L6a+* may have multiple replicas, it produces is a multiset (i.e., each bit can have multiple occurrences). Therefore, the effect will be proportional to the aggregate knowledge of all replicas in *L6a+*. If the total number of predictions in an output bit is below a certain threshold, the feedback value for that bit can be ignored. These modulatory inputs can arrive out-of-sync with the CC input activity (i.e., when the *L4* input arrive, or the EOS occurs). For example, *L1* feedback will arrive early in the sequence, while inhibition occurs at the end. Higher up in the hierarchy, the time between these two events can be substantial. Hence, a certain degree of permanence in the modulatory effect is required.

The proposed model for the cortico-cortical feedback is inspired by the effects of metabotropic channels on modulatory synapses [56] and how L6a→L4 intrinsical cortical projection modulates inhibitory outcomes in L4 [53]. We hypothesize that such effects on the target neuron are so small (weak postsynaptic depolarization of synapses located at the tip of the dendrite) that they only contribute to tilting the inhibition when the polarization of the competing dendrites is similar. Finally, the connections in our model lack plasticity. Therefore, modulatory inputs of *L4* should access any potentially active bit in the input. This is consistent with the extensive connectivity (>40%) observed in L6a→L4 projections [135]. Since L6a→L4 projections are predominantly inhibitory, to be consistent with our model, each axon might project to the non-corresponding L4 cells, as seems to occur in the TRN [14]. Each L6 axon efference copy excites multiple inhibitory cells in the TRN, inhibiting non-corresponding relay cells in the associated thalamic nuclei. Notwithstanding, we could not find any evidence to support that L6a+→L1 projections should match (the descending modulatory L6a+ axon should affect the neuron of the corresponding ascending driver axon). Very weak experimental evidence for this is presented in [76], where it is shown that for the case of the mouse's visual cortex, L6a→L1 are mostly monosynaptic and L6a→L4 polysynaptic. This suggests there is a mechanism involved in the neurodevelopment process that guides axonal targeting. Inhibitory late-spiking neurons in L1 [65], which mainly affect the L23 pyramidal cell apical tuft [68] may be responsible for this task. In our model the feedback modulation does not involve learning because we perfectly match the L6+ expectations with the pooling result. In a biological system, learning in that circuit may produce similar effects. Note that this seems to be consistent with the top-down hypothesis elucidated in L1 [66].

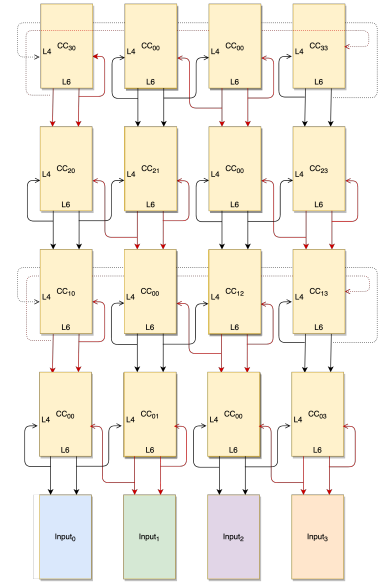

**Figure 10 Feedback Connections for Cortex in Figure 7**

Thus, the interpretation of feedback in our model differs significantly from the conventional view, which is derived from control theory and cybernetics [136]. From our point of view, interpreting the feedback as an "error signal" is misleading. Perhaps, this interpretation has been driven because a disturbance on the feedback signal will interrupt the CTLoop, which will have experimentally notorious consequences. The need to reconcile the DL learning mechanism with certain biological principles has also supported this vision [137]. An alternative interpretation such as "context signal" seems more appropriate. Although there are more sophisticated models, such as [138], they focus only on the behavioral interpretation of the feedback signal. Unlike our model, there is no clear connection to the underlying physiology.

---

[7] In absence of feedback, a stable sort is performed (e.g., by using a fixed tie breaker). Random tie breaking during the sorting process can make the system unstable. In biological systems morphological heterogeneity in synapses may have a similar effect.

### D. Supervised Learning (and the Matrix of the Thalamus)

Feedback interpretation can be useful to control the learning process. By externally modulating the L1 k-WTA result, we can manipulate the output sequence representation to achieve a desired value. Thus, it may be possible to implement some form of supervised learning.

In our system, this steering signal can be forced externally according to a learning goal. For example, this could be used to form invariant representations of multiple sequences (i.e., forcing a match of the *L1* pooling of different input sequences). In a biological system, steering signals may be generated internally via the corticostriatal loops[8]. Once the sequence is learned by the next rung, *L6a+* will adapt to the initial target signal, eliminating the need to maintain the supervision signal active.

This model is loosely inspired by the thalamic back-end of the Corticostriatal loops (CSLoop) [67][14]. For example, to align the input, the central pattern generators (i.e., the L1 output of the motor cortices) can be aligned with desired motor actions using the oculomotor loop. The cells in the matrix of the thalamus project diffusely over pia of the cortex[139]. This projection is like intracortical feedback. In addition, if the L5 output (i.e., sequence prediction) in motor cortex is understood as a the "motor" command, then by guiding L1 pooling, we are indirectly influencing the L5 output (since in our model L5 is only a predictor of the L1 pooling result).

This hypothesis is weakly supported by the fact that the matrix of the thalamus drives certain inhibitory cells in the L1 interneurons and the apical tuft of L23 pyramidal cells [69]. As hypothesized at the end of Section 0, L1 interneurons driven by L6b activity may be involved in the generation of the EOS. Then it seems plausible that if both the matrix and the feedback drive late-spiking interneurons in L1, both will have a similar effect on breaking the ties in the k-WTA used in the symbol generator (L1). In addition, the matrix affects non-late spiking L1 interneurons. Since we speculatively hypothesize that they are involved in EOS, we can extend this hypothesis to support externally generated EOS, as our model suggests. As shown in Figure 5, L1 can only learn if the CA3 above is present, therefore symbol sculpting ceases to affect the CC. Similarly, we can also disable externally induced EOS by BG. Therefore, the effects considered will only be noticeable when the CC is acquiring new knowledge.

The system in this work, only partially explore this idea implementing "externally generated EOS". It can be understood as a first order approximation to matrix-cortex connectivity. Although speculating with the CSLoop complexities is beyond the scope of this work, it could be used as an algorithmic mechanism to implement supervised learning. Therefore, by using specific cortex L5 outputs and behavioral targets to form the steering signal, it could be possible to guide the learning or cortical output, as seems to be occurring in biological systems [140]. Note that while this interpretation is consistent with the feedback purpose hypothesis of our model, it contradicts the current main hypothesis about the functionality of the thalamic matrix [141], [142].

### E. Cortical Column Scale-out and Output Interpretation

One of the key findings of the proposal is the limited variability of the CC output symbols. Our simulation-based evidence suggests that the system will not scale and will be brittle without adhering to the Reduced Number of Symbols Constraint principle (RNSC). Although the combinatorial nature can be used to establish a causal relationship between the sensory input and the cortical output (i.e., combining the symbols generated in all CCs at the last rung), when the signal is complex it may be difficult to correlate the sensor input to the output (e.g., to perform a classification task). An extreme manifestation of this problem arises when the output symbols in a CC are very close or identical, which is plausible due to the stochastic nature of the initial connectivity of *L1* and *L4*. In such a circumstance, the CCs in the lower rungs may cease to generate output symbols entirely, and certain CCs in the rung above could become inactive. Increasing the number of inputs into the cortex or replicating the internal modules of the CC may help reduce the likelihood of this issue. However, the possibility of having a silent CC in the last rung is not negligible.

---

[8] If the input is well known by the CC, these signals can be useful in helping the regular feedback to stabilize the output of the CC. This is consistent with the dual purpose of corticostriatal loops [162].

Nevertheless, it is possible to exploit the stochastic nature of initial connectivity and learning to increase cortical representational capacity. By using the output of a cortical column $CC_n^0$ into $n$ cortical columns in the next rung $CC_{n+1}^{0...n}$, which might find different representations for the data. Each $CC_{n+1}^{0...n}$ will react in a different way to the input data, depending on how the learning process and the projections are performed by *L4*, *L1* and the previous knowledge available in *L23*. Intuitively, each copy perceive the input projection from a slightly different perspective.

Hence, $CC_{n+1}^{0...n}$ will be building a different representation of the input data. Figure 11 exemplifies how that expansion might be done. In this case each CC is connected to two CC in the next rung (i.e., *n=2*). Therefore, the last rung will have 32 CC (4x8). Then, even if each $CC_{3x}$ generates a limited number of values the total number of combinations will be substantial. In general, $CC^{0..7}_{30..3}$ will generate a symbol out-of-sync. Hence, to classify an element, we should consider the outputs and the time interleaving during sensory exposure. At first glance internal replication in the CC seems to achieve similar effects. However, temporal coding is lost (since all replicas operate in synchrony). The proposed scale-out increases both symbol and temporal diversity and can be tuned according to the nature of the sensory stream (without changing the internal CC architecture). Additionally, this idea also helps stabilize CTLoop by enabling us to match each *L5* prediction with all the *L6a+* predictions from the upstream connected CCs. For example, we can improve the comparison's accuracy by applying an "and" operation to all *L6a+* predictions. Therefore, this idea can avoid us having to do a rung dependent CC scale-in adjustment to improve stability in the upper regions of the hierarchy. We can use a CC of fixed size and scale-out the rung size using cortical expansion. This, besides being more biologically plausible (since it removes the need for cellular specialization across regions), makes the cortex parameterization simpler and easier to adjust.

If we replicate CA3 slices independently, this architectural feature does not require changes to the internal CC structures. For example, $CC^0_{30}$ and $CC^0_{31}$ use a CA3 slice independent of $CC^1_{30}$ and $CC^2_{30}$. Therefore, the scalability of the system remains unchanged since there is no need to increase the size of the internal modules of the CC or the peripheral circuitry. This form of replication will complement the internal replication of the CC, because the feedback will be more stable. In addition, it will increase the resilience of the system to noise, since if a CC stops working all together, the replicas can still be used to generate output for the next rung.

According to this interpretation, the so-called "neural ensemble" is just a particular symbol produced by the EOS happening simultaneously in diverse CCs in the same rung. Thus, neither a firing neuron [143] nor the ensemble [144] has inherent meaning. Only the order in which the ensemble appears in a sequence is meaningful. In other words, the same ensemble can appear in completely unrelated events. In the same vein, using L5 to drive a hypothetical motor system will encode its output not as a single symbol but as a sequence of a reduced number of symbols to be interpreted by the peripheral circuits. Therefore, any system using final or intermediate outputs of the cortex should use the symbol sequence (e.g., cortex of the cortex for a classification system) not individual symbols. In this way, using pervasively a limited number of symbols does not compromise the pattern recognition capabilities of the system. This view may be understood as a hybrid between rate coding [145] and ensemble coding [144] (i.e., we form rates from a few different symbols). Figure 16 presents a graphical representation of this discussion.

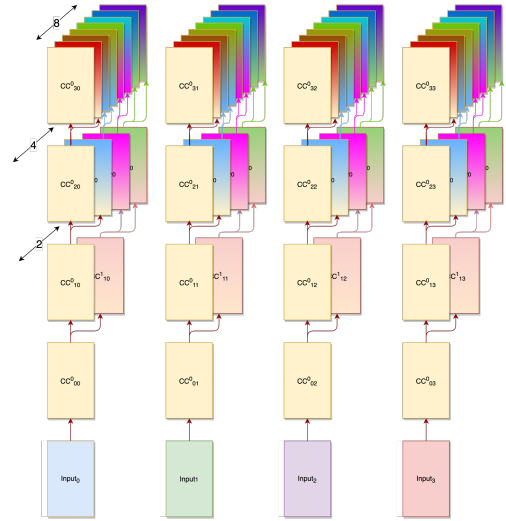

**Figure 11 Cortical Column Scale-out for cortex in Figure 7**

VI. HER AS AN AUTOMATIC SPEECH RECOGNITION SYSTEM

To examine the behavior of HER, we focus on its application to speech recognition. Next, we present an implementation for the auditory pathway and the cortical organization discussed earlier and assess the feasibility of a practical system.

A. *Cochlear Modeling and Automatic Input Gain*

A Mel spectrogram is computed from the audio signal. Unlike other ML approaches, the possible values of the features are binary: either the energy of the corresponding mel band is above a certain threshold ("1"), or it is below ("0"). This decision is required by the RNSC principle. Therefore, it is expected that the sensory systems (through brainstem circuitry) will also provide a very limited number of values as cortical inputs. The Distributed Representation (DR) used: "0" will be encoded as fixed small set of bits to 1 (e.g., 10 of 121 bits) and "1" is encoded as a larger set of bits set to 1 (e.g., 40 of 121 bits). Therefore, the coding is intensity-proportional and can only distinguish between two levels.

The system will discard fixed values in the signal (above a certain number of repetitions), so that only significant changes in the input signal are passed on to the cortex. Under such conditions, high temporal resolution is required to have a functional system. In contrast to conventional ML, HER uses a slide in the Fast-Fourier Transform (FFT) window of only 1ms (instead of 10-12.5ms). We will keep using the remaining FFT parameters similar (e.g., window width, filtering, etc..). This trade-off sacrifices magnitude precision for very precise timing. Finding this counterintuitive notion required considerable effort, but it is crucial and consistent with the fact that auditory nerve activity is highly unreliable[146]. Therefore, we hypothesize that biological systems can only be capable of considering an all-or-nothing situation.

A crucial aspect in this basic model is determining the threshold for each frequency, which is equivalent to controlling the input gain. Inspired by the descending reflex pathway of the auditory system, we modeled the medial olivocochlear reflex (MOCR)[147] to control the input gain. To accomplish this, we modeled a portion of the posterior ventral cochlear nucleus (pVCN). Our hypothesis is that this nucleus is responsible for continuously adjusting the threshold at each frequency to achieve a certain target average cochlear activity over time. The MOCR has two components: fast (~25 ms) and slow (~25 seconds) [148]. The first component is disregarded as it is believed to serve the purpose of safeguarding the internal structures of the cochlea from sudden increases in sound pressure. The slow component is the focus of our attention, and we approximate the cortical activity as a linear relation with the sensitivity threshold. The threshold of each band changes according to the following formula:

$$thr_{new} = \begin{cases} thr_{old}\left(1 + \left(\frac{ACT_{NEW} - ACT_{TUP}}{ACT_{TUP}}\right)k_{step}\right) \\ \quad if\ ACT_{NEW} > ACT_{TUP} \\ thr_{old}\left(1 - \left(\frac{ACT_{NEW} - ACT_{TDOWN}}{ACT_{TDOWN}}\right)k_{step}\right) \\ \quad if\ ACT_{NEW} < ACT_{TDOWN} \end{cases}$$

$$ACT_{NEW} = ACT_{NOW} \cdot \alpha + (1-\alpha) \cdot ACT_{OLD}$$

The change is controlled by a small adjustment in the threshold (e.g., $k_{step}$=4·$10^{-5}$, assuming that the value is updated every millisecond, which is equivalent to saying that it will take ~25 seconds to reach the desired activity). Accordingly, the new activity is estimated using an EMA of the previous 25 seconds. As a result, it will be possible to see small changes in threshold over long periods of the input signal. To ensure stability of the threshold, a hysteresis cycle is employed in the target activity (i.e., two margins in the target activity: $ACT_{TUP}$ and $ACT_{TDOWN}$). Therefore, if $ACT_{NEW}$ is above $ACT_{TUP}$ the threshold is lowered and if it is below $ACT_{TUP}$ it is raised. The threshold will be stabilized after a few seconds of signal. Using a faster accommodation result in context-dependent coding, which is incorrect. The $ACT_{TUP}$ and $ACT_{TDOWN}$ are set to minimize changes in the cochlear output (for a 10-bit set SDR in "0" and 40-bits set in "1" to values between ~13-14). Thus, our intuition is that the goal of the MOCR, as well as protecting the cochlear structures from damage, is to provide the minimum number of "1s" to allow the cortex to learn the model carried by the signal. Thus, the MOCR is likely to have a significant impact on high-level constructs, which is consistent with some empirical evidence [149].

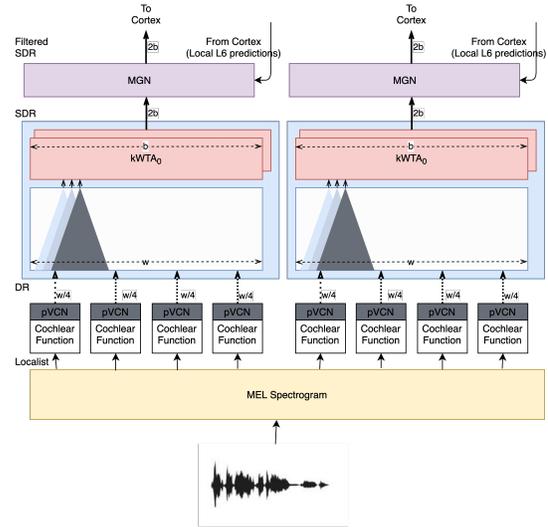

Figure 12 Encoding Process (for a 2xn Cortex)

### B. *Sparsification and Cortical Input Adaptation*

Cochlear output does not obey the sparsity rules or may not provide useful information to the cortex. We introduce two additional components inspired in the Dorsal Cochlear Nucleus (DCN) and the Medial Geniculate Nucleus (MGN) that focus on performing sparsification and discarding non-useful cochlear output.

In *DCN*, which is loosely inspired in *L4*, we will compute a LSH of the cochlear output via a sparse (e.g., 2%) k-WTA inhibition of the auditory nerve projections of a subset the of signal features. In contrast with *L4* where the receptive field is module-wide with a ~20% connectivity wide receptive field, DCN parameters have more resemblance with the retina: reduced receptive field (~10%) and full connectivity. For a cortex of *n* width (i.e., *n*-cortical columns), and *f* features of the speech signal, we combine tonotopically arrangement *n/f* features per DCN component. The proximal synapses of the output cells are set at boot time with a receptive field of size *n/f·b b* being the number of output bits of the component. The synapses have no plasticity. Hence in the time step the DR representation of the input signal is transformed into an SDR. Figure 12 shows a sketch of the DCN functionality.

MGN additionally performs sensory adaption functions by discarding non-useful inputs. If a part of the DCN module is providing a constant value during a sufficiently long time, the input to corresponding CCs is suppressed. The rationale is to spare synaptic load in the Sequence Memories of the CC. Although it is important to disambiguate fixed short signals, to retain the capability to disambiguate very long fixed signals requires additional contexts for the same value (i.e., more branches). This is highly taxing from a synaptic load perspective and has diminishing returns. Intuitively this can be seen as a form of "edge detection" in the corresponding power band of the signal.

The sensory adaptation function is inspired in the triadic synapses [56]. Such a circuit is composed of a glomerulus of three components: retinal ganglion cell terminal and dendrites of an inhibitory and relay cell from LGN. High activity in the axonal terminal will activate, via metabotropic synapses, the dendrite of the inhibitory cell, which will silence the dendritic terminal of the relay cell. Our hypothesis is that such a mechanism will attenuate input afferents that are used repeatedly (i.e., there is a fixed signal in the sensor). We speculatively assume that similar mechanism is used not only by any first order thalamic nuclei, such as MGBv, but also in high order nuclei, such as MGBc [150]. This consideration is assumed by looking at experimental evidence such as [151]. Therefore, in CTLoop we limit the number of fixed forward predictions accordingly. Perhaps not coincidentally, not performing this cancellation introduces the possibility of the appearance of prediction runaways, in which L5 and L6+ enter a never-ending loop of constant values. Most likely, biological systems will use additional mechanisms in MGN [152], given the system we have we restrict our model to this.

Finally, we incorporate the idea depicted in Section III.D to the MGN. If there is no CA3 slice allocated above the MGN we will suppress any input bits coming from DCN that are not present in *L6+* prediction.

### C. Brainstem Biological Relevance in the Cortical Input/Output

Most likely, biological systems have a complexity that is not considered here: L5 tonotopic outputs affect the DCN via parallel fibers and learning on distal segments of fusiform cells (FC) and a network of inhibitory networks around cart-wheel cells (CWC). Both FC and CWC distal cells have a morphology similar to cerebellar Purkinje cells [153]. Here, we model only the proximal dendrites of the FC, which actually have no learning capacity [15], and the tuberculoventral cell, which may be responsible for carrying out the inhibition process in the FC. In addition to the CWC, many other classes of cells in the DCN are not included in the model.

We hypothesize that biological systems utilize L5 predictions, when available (i.e., when attentional mechanisms are engaged), to filter out instabilities in the auditory nerve. Note that in addition to L5 information, other sensory information, such as that coming from the vestibular system, is compounded in the inputs of the distal dendrites of FC and CWC.

We speculatively hypothesize that a significant portion of brainstem nuclei (and the so-called motor afferents) in conjunction with the cerebellum is dedicated to stabilizing cortical input (and output). The system will use pervasive attention mechanism through L5 predictions to stabilize sensory information in a very sparse set of inputs and output values (or symbols). However, given the complexity of the matter, we have chosen here to keep such issues out of the model. By using a stable input at the cochlea, the auditory nerve will always be consistent for the same input signal (since we are using a noise free synthetic system). A noisy signal, an unstable cochlea, or an inhibition process in the DCN may need to be reconsidered in future work. Although MOCR filters out some of the noise in the signal, such a complex scenario will most likely require detailed modeling of the DCN. In any case, it should be noted that the proposed model is consistent with this view.

## VII. EVALUATION

### A. Practical Implementations of the System

An in-house C++ simulator, called CTXSim was developed [154]. To cope with the large number of components, the simulator may exploit thread level parallelism. Since the communication between components is sparse (compared to the computational cost of simulating the internals of the CC), the synchronization overhead is low. Here we will only use shared memory implementation. In any case, note that the activity rate is dominated by lower rungs of the cortex. Higher in the hierarchy the dimensional reduction is large enough to make its contribution to the total computational cost negligible. On a modern server, a 1x5 cortex running in a single core achieves a speedup of approximately 100 times over real-time audio. The low computational cost of prediction (essentially comparisons) and learning (low precision subtractions and additions) makes current GPU or TPU architectures unattractive. However, it is reasonable to assume that scaling the system size to hundreds or thousands of CCs is feasible using Parallel Discrete Event Simulation paradigms, not only in many-core systems but also in large-scale clusters. This is due to the sparse communication between CCs. By increasing the replication level, it may be possible to use pervasively optimistic concurrency without affecting the system behavior significantly. Further optimizations, not explored here, included hand-optimized code using SPMD ISA extensions, which could significantly increase single core performance. We have only used a C++ compiler capable of emitting code optimized for the family of processors used by the servers.

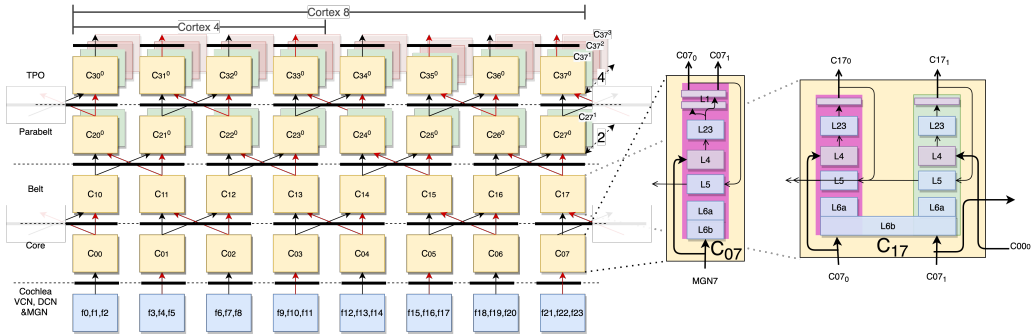

**Figure 13 Architecture of the Simulated Cortex**
**(12/24 frequency bands, 4/8x4x{1,1,2,4} CC), CC details for CCs in the first and subsequent rungs**

Adapting existing HTM hardware implementations, such as [155], for this proposal, a frugal large-scale system can be built. These systems, with a mixed signal design, using a CLA with 960 bits and 256 dendritic segments per bit, is capable of operating with less with ~30mW at ~100MHz. A Cortical Column using 121-bit replicas, 4 replicas in L23/L6, and 2 Forward Predictors in L5, is equivalent to a ~1200-bit CLA system (121·(4·2)+121·2). Therefore, assuming linear power scaling and 10mW for temporal pooling, a Cortical Column will require ~50mW. Considering a square cortex *NxN* with a given power per Cortical Column *P0*, a fixed dimensional reduction between layers of *a*, the total power budget *Pt* can be expressed as: $Pt = P_0 \cdot N \cdot \sum_{i=0}^{N-1} a^i$. With a fixed *a=1/10*, solving *N* in this equation for a power budget of 20Watts, assuming that inter-chip communication is sparse enough to have a negligible power overhead, it could be feasible to have a cortex with 64x64 Cortical Columns. In addition, HER frugality suggests that it may be amenable to 3D Stacked Processing-in-memory systems [156][157], which could enable biological-level scale systems.

### B. *Input Dataset*

The input data stream consist of two parts: (1) One generated using Spanish sentences (~700 sentences with 1500 different words) taken from the *Shardvard-corpus* [158]. The audio files are generated using *Tacotron* [159] for a single speaker model, trained over 400K iterations. (2) The data used to train the model, consisting of three available audiobooks with ~11,000 non-synthetic long sentences of the same speaker [160]. Since this is a proof of concept, we will not compare its accuracy with another ASR. Furthermore, the working conditions considered here cannot be handled by any known ML system. Unlike other ASR systems, here we start with a completely empty system, with a reduced amount of unlabeled data, so the raw comparison will be unfair (note that state-of-the-art ASR systems require terabytes of speech and millions of GPU hours of training [161]). In a way, our system starts as a *tabula rasa*[9]: no-knowledge and no supervised learning (at this point). The objective of this evaluation is not to develop a system ready for production, but rather to assess the feasibility of the proposed approach for addressing a challenging problem like speech recognition.

### C. *System configuration*

Figure 13 shows the architecture of the cortex, which is loosely inspired by the auditory cortex of biological systems. Although cortical homogeneity has been debunked [162], experimental evidence such as [16] still remains puzzling. Here we assume this scenario and consider the internal architecture of the CC to be uniform. This simplifies the search for architectural parameters and enhances the elegance of the design. The system has either 12 or 24 CCs using a lateral contextualization of a single CC in third and fourth levels. The number of CC will grow in upper levels to extend representational capabilities. The proposed scale-out mechanism introduced in Figure 11, is utilized to expand the third and the third and fouth rungs to 2 and 4 CCs, respectively.

To mantain all CCs symmetrical, the lateral contextualization uses wraparounds. CA3 slices are synchronizing learning advances in groups of two CCs. The CA3 are shifted to align the learning status through rungs. CA3 slices are placed at the cortical exit to facilitate attention downstream. The CC sizes are tuned to a single module in the first levels. Given the reliable substrate with a perfect synaptic transmission scenario such as this, the use of additional modules does not appear to provide benefits, while increasing the computational cost. For the same reasons (plus a significant increase in the synaptic load of the lowest rung, which contradicts experimental evidence [2]), there is no lateral contextualization in the first level. Starting from rung 2 each CC has two modules, of which one is vertical, and one is lateral. Each module, inspired by[123], has 121 cells. In all rungs except the first and second, every pyramidal cell in *L23*, *L5*, and *L6a/b* has 16 dendritic branches. In the first and second rungs, there are only 4 and 8 branches, respectively. There is no limit to the number of dendritic segments per branch. The dendritic segments have up to 128 synapses. To minimize the computational cost, none of the modules have internal replicas. Figure 13 on the left illustrates the CC in the first and subsequent rungs.

---

[9] While parameters and cortex architecture are predefined, active and silent (proximal) synapses in L4 and L1 are randomly created at boot time. All CA3/L5/L23/L6 have no (distal) synapses at boot time.

The system is highly resilient to most parameters, making it difficult to identify when it was not functioning properly during the design phase. Identifying whether the issue was caused by the algorithm, the parameters, or the implementation, was challenging. However, once all components of the architecture were in place, the sensitivity to most parameters was significantly reduced. In any case, parameters can be adjusted for other types of sensory information. In biological systems, the anatomy of the cortex (and therefore its parametrization) is not homogenous [163]. It is unclear to us whether this is a consequence of epigenic effects (during the developmental process or postnatally) or a result of the evolutionary process.

Although not explored here, with more complex inputs it seems possible to optimize the parameters using some form of genetic algorithm. The cortex may also be rearchitected through this kind of approaches, especially if the system and the input are complex (e.g., volitional data streams may require it). Therefore, the system parameterization can be understood as an innate meta-parameter that can be encoded as genetic information in biological systems [164]. Since it is possible to consider what impact a parameter might have on the system behavior, the selection can be guided.

Given the limited constrained complexity of the problem at hand, we have used experimental evidence on cortex and speech as the main guides to manually tune the learning parameters. The objective of the configuration is to reduce the number of synapses as much as possible while maintaining disambiguation capabilities and stability. It appears that exploiting the hierarchical organization is crucial to reconcile these three requirements. It cannot be claimed that the chosen parameters are optimal: better combinations might be possible. Additionally, certain parameters that are considered fixed may be dependent on activity.

To handle the large volume of ongoing new sequences, the learning rate in the distal synapses of the entire cortical column should be small, approximately $\sim 10^{-5}$). The system is assumed to be immersed in an uncontrolled learning process (i.e., exposed to the entire data set at once). Under such circumstances, the learning modulation is key to the orderly acquisition of the abundant novel incoming sensory data.

According to the RNSC principle, the system uses a counterintuitive key observation: *instantaneous accuracy is irrelevant*. What is important is the spatial (i.e., neighboring sensor or CC) and temporal context (i.e., previous values in the time series) of the value. Although the system can operate with high resolution in the input stream, it will not be able to reach a steady state with complex input data. At high precision, the system requires a large synaptic load in the first rung with many long sequences. This is contrary to the fundamentals of hierarchical memory organization. In fact, it is well known that in biological systems the primary sensory cortices' pyramidal cells are so sparse that they are called granular cortex [108] (due to the prominence of granular cells). After a considerable amount of work, we come to the surprising conclusion that the system, as designed, can learn only with two buckets per audio feature. This, as discussed in Section VI.A., allows us to reduce the cortical input activity and easily self-regulate the gain of the encoders. Using one bit per feature allows us to combine three features per encoder. Hence in an 8x4 we can process 24 audio features, which may be sufficient for speech recognition tasks.

The most important set of parameters are the rolling windows of the exponential moving averages on different Sequence Memories. These determine the average sequence length. If we consider the highly speculative hypothesis that LFP is epiphenomenal to event segmentation, we can look for hints in different wavelengths of system-level oscillations in the brain.

Inspired by the replay length sequence in the hippocampal complex [165], we chose a large rolling window in CA3 slices with a narrow margin in *Up* and *Down* parameters (alfa=0.01, Down=0.05, Up=0.3). The recurrence delay in replay is set according to the theta oscillation frequency characteristic of SWR, which ranges from approximately 4Hz to 12Hz. The hysteresis's alfa parameter (or *time constant*) is kept constant from rung to rung, resulting in a progressive increase in the rolling window. As previously discussed, CA3 should prevent incomplete identification from progressing throughout the hierarchy.

The parameters of the learning hysteresis cycle in L6b are configured for rapid action. This allows the input stream to be segmented into shorter sequences, which is advantageous for utilizing hierarchical memory organization (alfa=0.9, Down=0.1, Up=0.5). Intuitively, the "usefulness" of a synapse at the lowest levels (i.e., the probability of being involved in the identification of a given input) is higher when the sequences are short. L23 and L6a use medium timing constants (alfa=0.1, Down=0.05, Up=0.4) to filter out spurious data with the hysteresis cycle while maintaining long-term stability. L5 is set between L6a and L6b (alfa=0.5, Down=0.05, Up=0.5). The objective is to reduce the length of prediction sequences to increase the chances of having a precise enough match in the CTLoop while maintaining stability. Note that L23, L6a, and L5 are kept learning throughout the entire system lifespan. Learning is disabled in L6b only when the CA3 above and below are unallocated to prevent learning noise due to its large alfa. The combination of two observations from [166] weakly support L6b parameters: (1) V1 oscillations are in the 60-80Hz range and (2) V2 in the 14-18Hz range (i.e., 4 times less activity). Hence, we can speculatively hypothesize that the dimensional reduction across rungs is approximately four times. The fixed alfa value across rungs can be emulated with a variable interarrival time exponential moving average estimation and a variable rolling window estimated using dimensional reduction (or the other way around). If biological systems have some mechanism for self-regulating how much of the past is considered, our model may be plausible. However, we could not find any evidence of this in the literature.

In all layers, the LTD to LTP ratio depends on the alfa used in the hysteresis cycle. In any sequence memory, the ratio is set to optimize the time needed to segregate all the contexts of subsequences in the same sequence (i.e., to isolate the dendritic branches used in each one). For alfa 0.1 (CA3, L23, L6a), it was found that a ratio above 10% may be excessively harsh. For L6b and L5, ratios of 70% and 50% are tolerated while improving segmentation homogeneity and CTC loop forwarding, respectively.

LTD induced by Hetero-synaptic plasticity [167] is only utilized in L1 and L4, with a 10% ratio. In L23, L6a/b, and L5, causes instability when different sequences share a dendritic arbor. In L1 and L4, as discussed above, hetero-synaptic LTD is required to prune silent synapses. These values differ significantly with [18]: LTD are significantly larger, there is LTD in spatial pooling and there is no hetero-synaptic plasticity in temporal pooling. The learning rate in distal dendrites is set to $10^{-5}$ while the permanence ranges from 0.0 to 1.0. Long term modulation is homogeneous, being 100x (d) through all rungs. CA3 learning rate is 100x.

There is no short-term learning modulation in L1/L4. The learning rate is fast (10% of the maximum permanence) with 1% for heterosynaptic forgetting and LTD. A fast-learning rate is required to accelerate cluster formation of the output projections (that may not be biologically plausible). A higher value appears to decrease cortical disambiguation capabilities. Despite the significant learning rate, SWR remains beneficial. LTD is applied to all non-firing post synaptic cells with permanence above the threshold. LTD in proximal dendrites increases the compactness of the clusters. Synapses that reach zero permanence are pruned away (and never regrow). A synapse is pruned out when its permanence reaches zero. On average, it takes ~$10^9$ cycles to prune away an unused proximal synapse. In general, the system is stable with a high LTD/LTP ratio and both implicit and explicit synaptic pruning. We hypothesize that in symbol formation (at any level) such a ratio maximizes the distance between symbols, reducing the effectiveness of the system under the RCNS principle. All (hyper) parameters of the system are detailed in [154].

VIII. RESULTS I: COMPONENTS OF HER

A. *Hierarchical Continuous Learning*

In this initial experiment, we will demonstrate that the system's behavior aligns with expectations. Specifically, we will show how: (1) it reduces the dimensionality of the data across rungs, (2) it is able to do so both from a blank slate or with new data added to previously acquired knowledge. Learning is performed in open loop (i.e., no supervision in the segmentations is done), the SWR mechanism is not in use, and the CTLoop is unavailable. A cortex with twelve frequency bands (4x4x{1,1,2,4}) is used.

To train the system for (1), we use the first 10 sentences of the synthetic corpus in sequential order repeatedly. Once the learning is done at all rungs, we reset the system ongoing predictions and expose it again to the same input. The results are presented in Figure 14, which includes the raw audio signal and the spectrogram of the first 10 seconds at the bottom. The figure displays the exponential moving average of the anomaly score (EMAAS) of L6b in the first four CCs of the first rung, with sudden changes in AS indicating the end of sequence (EOS) detected from the data at the cortical input. The upper part of the figure presents the EMAAS of the CCs from the second to the fourth level. Zooming out to 100 and 500 seconds reveals the periodicity of the signal. Only the first out-scale CC in level three and four is shown. As we ascend, there is a noticeable decrease in activity, specifically a significant reduction in the number of EOSs. Shortly after the initial moments, when the system state is cleared, it becomes evident that the system detects the periodicity in the signal. The 10 sentences take approximately 21 seconds. The input's periodicity becomes apparent in rung 4, where most CCs exhibit a frequency that is perfectly aligned with the 10-sentence duration (~21 seconds). The system acquired this knowledge through open-loop learning, without supervision as described in Section V.D. Therefore, data segmentation is a result of the stochastic learning process and sensory input from the prime state. The data is not accurately segmented according to phonemes, words, or sentences, although the periodicity is similar.

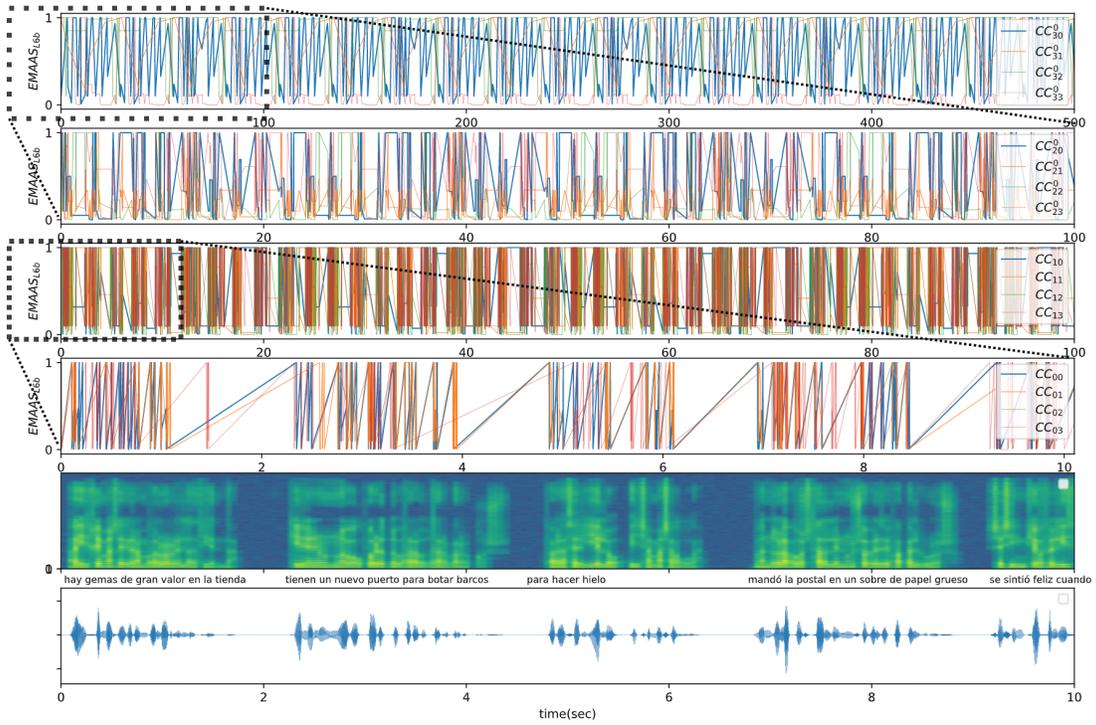

**Figure 14** Visual representation of dimensional reduction from raw data and spectrogram of ten sentences fed sequentially to End-of-Sequence identifications over 16 CC of the cortex (Exponential moving average of the Anomaly Score of L6b).

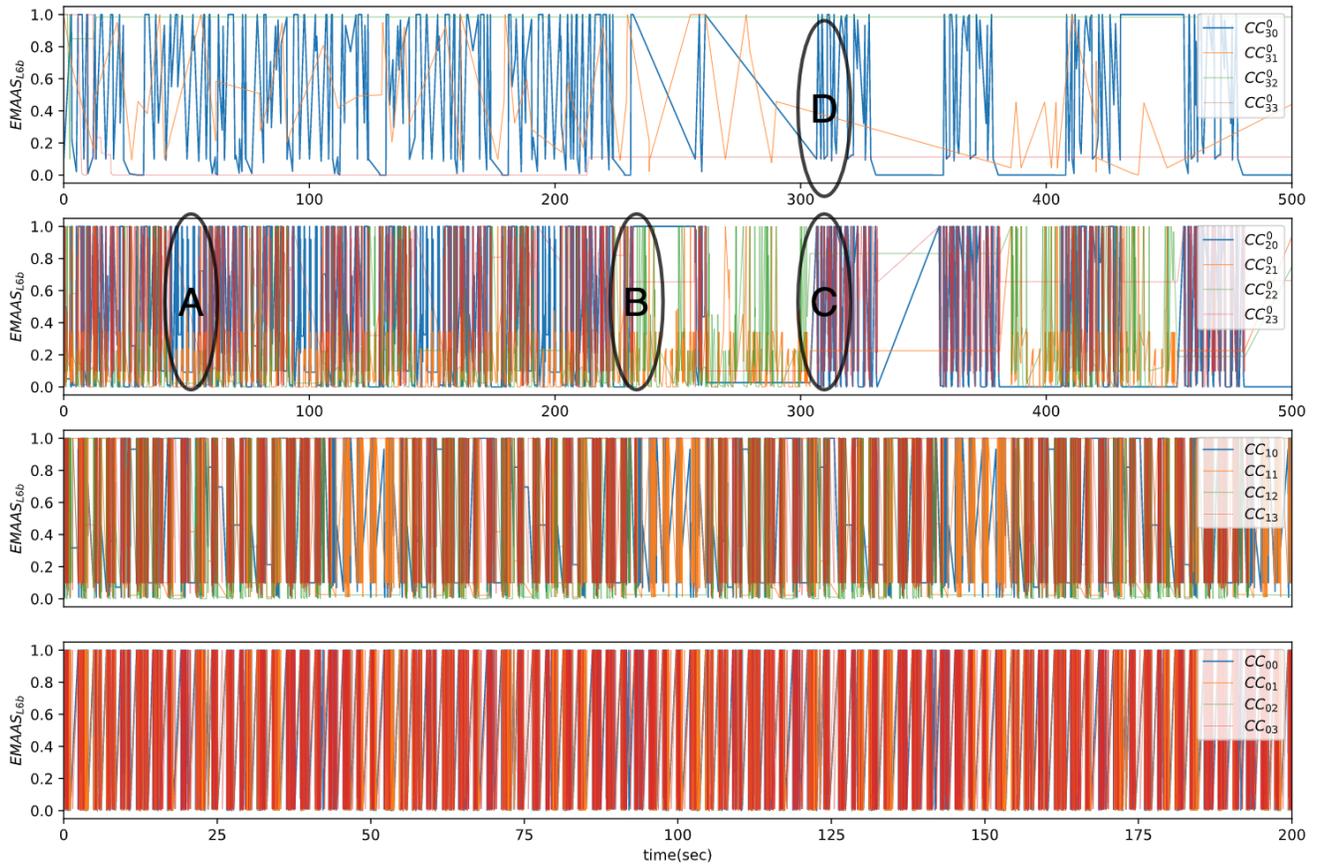

**Figure 15** L6b anomaly score when 10 new sentences are added to the input sequence.

Figure 15 illustrates the evolution of EMA L6b layers when repeatedly fed with a sequence of 20 sentences. The first 10 sentences are known, while the last 10 sentences are unknown. The first two rungs quickly accommodate the data. However, certain rung 3 CCs are unable to identify the new sentences in zone A. This allocation of CA3 disables learning in the affected CC to the point where the system response is interrupted (zone B), which also disrupts the corresponding CC in the next rung. However, the system can only recognize the known sentences (zone C), which partially provide data to the next rung (zone D). After the reactivation of CA3 slices, there is a break in the signal periodicity, and the known data is intermittently identified.

Figure 16 clarifies the discussion in Section V.E and the RNSC principle, by showing how the cortex reacts to the groups of sentences in this example (s1-s10). In Cortex 4 in Figure 13, which is expanded in (**A**), there are 16 CCs in the last rung. For each EOS in (**B**), one or more CC will be generating a symbol. This is done with the periodicity of the input. Focusing on the topmost group of CC in the left cortex column (i.e., $CC_{03}^{\{0..3\}}$), the number of symbols per CC ranges from 2 to 4, labeled as A00,B00, A01, etc.. (**C**). Similar symbol counts are found in the CCs of other cortex columns. Increasing the input dataset yields similar counts. Each number in the SDR represents a winner for each symbol, with inhibition on each L1. There are two L1 per CC, aligned with L23 replicas, each consisting of 121 cells. The sparsity is set to 4/121 (~3.3%) as per configuration. As shown, certain CCs have close symbols (e.g., B01 and B02), while others are more distant (e.g., A00 and A01). This is determined by the initial connectivity, data, and the stochastic learning process. The paragraph will be represented by a sequence of 4x4 ensembles with 1 to 16 symbols, as shown in (**D**).

The CTLoop only makes the interleaving relevant, i.e., the ordering of the ensembles, not the delay between them. In this case, the sequence has six ensembles, that can be flattened into one (**E**). The time-encoded version can be used if the cortex is connected to a time aware component, or the flattened version can be used if the cortex is connected to a conventional time-agnostic component (e.g., a MLP layer)[10]. When the same CC have multiple EOS, the symbols are combined in an array (e.g., $CC_{32}^{1}$ has three values). An alternative way to represent this flat mapping is to convert it into a feature vector of floats, with one dimension per cell. For example, in a Cortex-8, this would result in a sparse vector of approximately 7,700 dimensions ($8 \cdot 4 \cdot 121 \cdot 2$), with each component representing the normalized activation rate of the corresponding cell. This approach enables the combination of both code and rate in a sparse distributed representation of small integers. Such format can be useful for computing similarities or identifying sequences. This can be further distilled down to just a vector of rates with one dimension per CC. However, it is important to note that these representations ignore time interleaving and coding respectively, which may lead to false identifications.

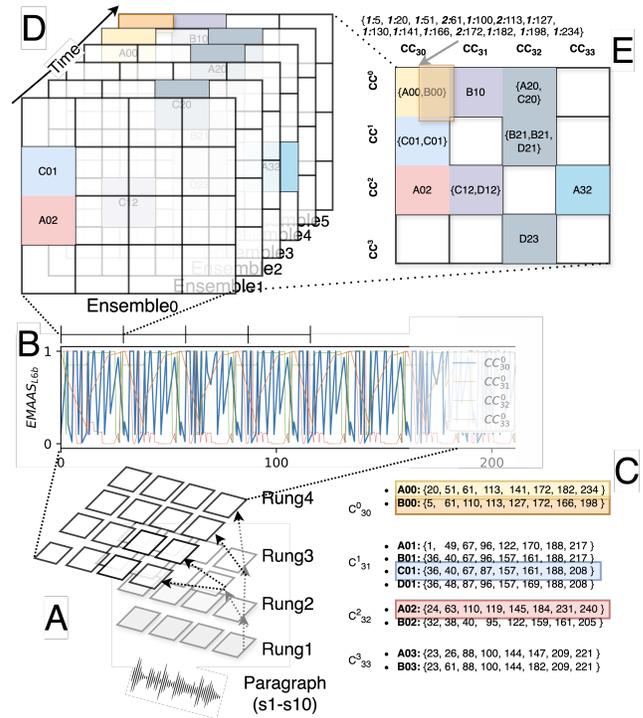

**Figure 16 Example of Cortex Output Interpretation**

Even with a reduced number of symbols per CC, the identification capabilities seem to be limitless. Therefore, the RNSC principle does not restrict cortical representation capabilities. The purpose of this paper is to showcase the proposal's characteristics. The identification capabilities can be utilized for a practical system. For instance, the same sentence encoding can be obtained at rung 3, word-alike encoding at rung 2, and phoneme-alike at the rung 1. Note that L5 output represent predicted next sequence. Extending the idea suggested in section V.D, may enable us to drive the system towards specific goals, but they are beyond the scope of the paper.

---

[10] Adapting this code as input for a system, such as a DL NN, may require some adjustments due to representational drift. Most likely it will require a system of similar nature to compensate such drift.

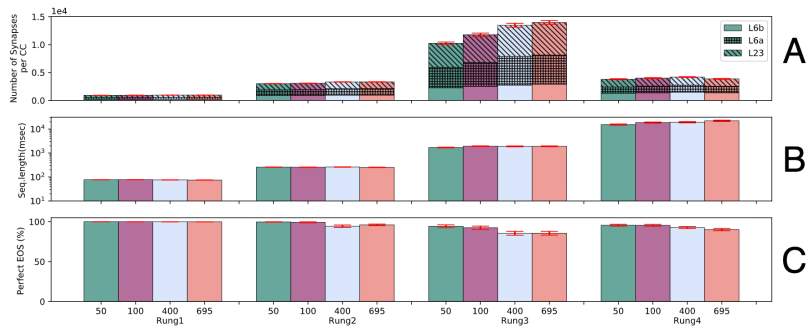

**Figure 17 Results for 50, 100, 500 and 695 sentences from Sharvard Corpus fed sequentially into the cortex:
A: Average number of Synapses per Cortical Column, B: Average Sequence Length, C: Percentage of Perfect End of Sequence**

### B. *Hierarchical Storage Scaling (and Measuring System Stability)*

To better understand the previous results, especially the response of the lowest rungs when an additional 10 sentences are added to the input, we will next expose the system (from the prime state) to an increasing number of sentences (10, 50, 100 and 500) from the *Sharvard* corpus with 24 audio features (Cotex8). All the sentences are exposed sequentially and repeatedly. Each experiment is conducted multiple times to ensure statistically significant results.

Before proceeding we need a metric to determine when the system has acquired the knowledge. Unlike conventional DL systems, there is no simple performance metric, such as classification accuracy or mean absolute error, as there is no learning target (at this time). As previously stated, analyzing the system stability by directly handling the cortex output encoding may prove to be a challenging task. Since the input is periodic, we introduce a metric that serves as a proxy for the stability of cortical state. At audio feature extraction each input is uniquely tagged with the sentence *id* and the window offset from the origin of the sentence. By propagating this tag, attached to the symbol generation, we can associate each EOS with the tag found in the "offending" input. Tracking such tags, we can determine for each sentence of the inputs how many occurrences of such sequence is found in each CC in the system. In a perfectly stable system for a given input, the number of occurrences for each tag should be equal in each period, which is called **Perfect EOS**. We perform this analysis periodically until we have enough samples. The Perfect EOS ratio will provide a good idea about how stable the system is, without requiring cortical output interpretation[11] (shown in Figure 16). The proposed metric formalizes the visual interpretation of Figure 14 and has, in fact, guided the system's design.

Figure 17 summarizes the results of this with the confidence intervals for a 99.5% likelihood and all CC. Simulations ends when all CC reach a Perfect EOS above 95%. Synaptic load includes the distal synapses of the Sequence Memories in use. Note that CTLoop is disabled, hence L5 is disabled.

Upon examination of the synaptic load, it becomes apparent that it remains constant in the first two rungs. This suggests that the system is identifying the "fundamental" components of speech and as expected, remains unchanged through the growing number of sentences. The effect is more noticeable in the first rung, with a slight increase in the second. Although this extraction depends on the actual data, how the system is exposed to the new data and the stochasticity of the learning algorithm, the variability is minimal. From the third rung upwards, the situation changes: there is a clear effect on the synaptic load of the number of sentences, reaching an aggregate synaptic load of 15,000 synapses per CC with 695 sentences. On average, the total synaptic load of the cortex is approximately 430K synapses.

The length of the sequences found in each input on the first three rungs are very similar. They are ~80±12ms, for rung1, ~250ms±6ms for rung2, from ~1.6sec to 1.95 sec for rung 3 and from ~15sec to ~22sec for rung4. The input activity at rung1 is ~125Hz. The triadic circuit in the MGN filters out approximately 87% of the sensory stream, resulting in a cortical input activity that is consistent with gamma frequency (30-100Hz). In addition to their similarity to beta (12-30Hz) and theta (0.5-4Hz), the timing of the sequences discovered by the first and second rung are remarkably close to the average phoneme and syllable length in

---

[11] In a completely stable system, and unlike time ensemble interleaving, symbol representations change occasionally (e.g., a few bits in an ensemble representation may be altered).

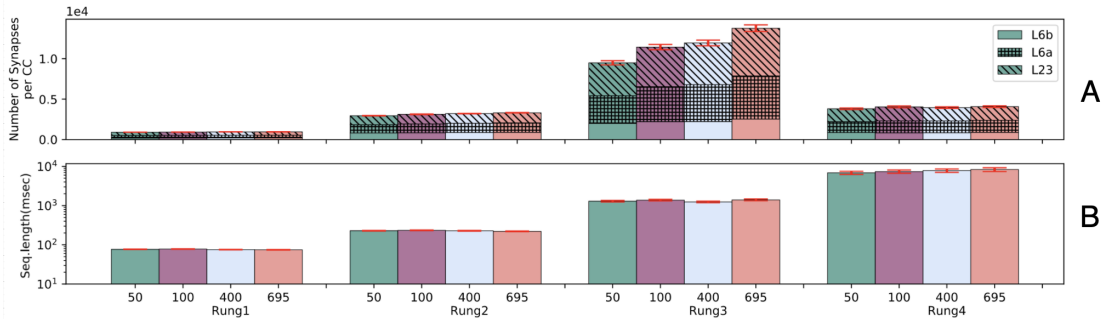
Figure 18 Results for 50, 100, 400 and 695 sentences from Sharvard Corpus fed sequentially, forcing EOS during learning:
A: Average synaptic load per cortical column, B: Average sequence length at each cortical rung.

Spanish [168]. Based on our interpretation of the CC output, the first rung is encoding phonemes as sequences of cochlear input. The second rung is encoding syllables as sequences of phonemes. The synaptic load becomes noticeable from the third rung, as the diversity of ensemble at the output (i.e., syllables) of the second rung is higher. In contrast, the ensemble diversity from the first is not, which is consistent with the limited number of phonemes in the Spanish language (24-27). The result is also consistent with the fact that speech in different languages appears to have a peak between 4 and 5Hz [169]. Perhaps not coincidentally, this is consistent with visual stimuli [170], or the processing of visual stimuli [171]. This may explain why in other biological systems that lack the neural machinery required to understand human language form similar constructs (i.e., oscillatory responses with a periodicity of about 300ms) in the primary auditory cortex [172].The result (and the language itself) is just a consequence of the input frequency and the dimensional reduction from the rung.

### C. Supervised Learning (Partial Corticostriatal Loop)

The sequence segmentation beyond the second rung is inconsistent as we increase the number of sentences. There is a bundling of sentences in the cortex is a result of the low diversity of words and the challenging sentence ordering. In a natural environment, there would be more diversity both in words, sentences, and ordering. In our scenario, the system is stochastically joining sentences and words because the syllables at the border always appear in the same order. This is particularly acute in the fourth rung with bundles more than 20 sentences per EOS. Under these conditions, the system's ability to disambiguate the input is reduced due to the high dimensional reduction from rung to rung, which exceeds four. To address this issue, the system requires some sort of learning supervision. Using the corticostriatal loops idea (See Section V.D) we will use a partial implementation as proof of concept. The loop only includes the ascending pathway. Currently, we do not have a model for either the architecture of the descending path or how to self-generate the signals that it carries. However, it is likely that agency will be required in the upper rungs of the cortex, which HER currently lacks. Nevertheless, by using the input tags we can identify any changes in the sentence ID to which the inputs belong. In any CC with the *L6b* learning active (i.e., either *CA3* above or below is allocated), we will forcibly insert an EOS. When the CC reaches the stable state, the mechanism stops acting. Therefore, we are only proposing a partial implementation. There is no modulation for the output symbol of the EOS during learning, nor any symbol stabilization assistance outside learning.

Figure 18 shows the impact of the mechanism for the same experiment. The sequence length remains unchanged in the two first rungs but becomes significantly more consistent in the third and fourth rungs as the number of sentences increases. Except for 10 sentences, where the total sequence may be too short, now the sequence length in the third and fourth rungs are 1.3±0.3seconds and 7.5±0.1seconds respectively. Since approximately half of the audio files correspond to sentence segmentation, this is more consistent with word duration (~650ms). At the fourth rung the system is still identifying groups of ~3 sentences. The synaptic load of the last two rungs is slightly higher due to this mechanism. Thanks to it, the system gains a higher symbol diversity of approximately 4 per CC, although this may vary from CC to CC and simulation to simulation. When combined with more extensive temporal activity, it allows for better disambiguation capabilities.

### D. Learning Convergence and Synaptic Load Improvement (Hippocampal Replay)

Synaptic load in the previous experiment shows, especially in the third rung, a progression in L23 synaptic load that seems more acute than in other layers. Since L23 learns from L4 projections, it seems reasonable to suppose that this layer does not converge the input projections fast enough. Similarly, L6a/b will be affected by L1 slow cluster formation. Next, we will demonstrate how adding the SWR feature to the cortex can alleviate the problem. To better understand the impact of SWR metrics Figure 19 compares the dynamic evolution of two cortices, one with the mechanism and one without. For the sake of clarity, only the input with 50 sequential sentences is shown. Other inputs demonstrate comparable behavior.

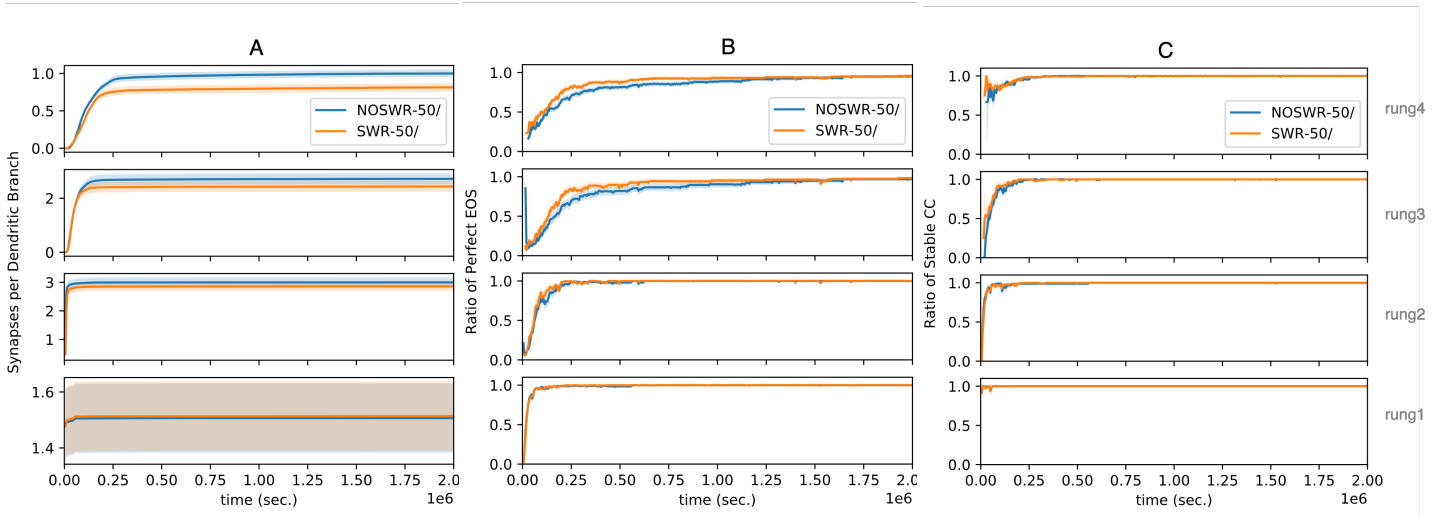

**Figure 19** Performance Metrics across Learning comparing a non-SWR and SWR cortices for 50 sentences from *Sharvard* exposed sequentially, A: Synaptic Load, B: Ratio of Perfect EOS, C: Ratio of cortical column stability over periods of 5000 seconds.

Figure 19.A shows the average number of distal synapses per dendritic branch[12] (DB) in all Sequence Memories of all CCs in each rung. The first rung has 4DB, the second 8DB and, the rest 16DB. Although this is done to optimize memory requirements of each simulation, the last rung in this sequential scenario is oversized. The shaded region in the figures represents the confidence interval for the value in each CC through all the simulations with a 99.5% likelihood.

In the lowest rung, there is no effect of SWR, since the shaded areas overlap. However, as we move up the hierarchy, there is a clear reduction in synaptic load, with the last rung showing a reduction of close to 20%. Beyond that saving, the cortex stabilizes sooner, as shown in Figure 19.B. The cortex with SWR reaches a near 100% perfect EOS significantly faster than the other. As with synaptic load, this is more clearly distinguishable in the third and fourth rungs, where SWR has a higher probability of occurrence.

It is important to note that Perfect EOS convergence occurs after synaptic load stabilization, which is indicated by horizontal lines. In certain usage scenarios, determining the Perfect EOS metric may not be easy. Figure 19.C introduces an alternative cortical knowledge metric that correlates with it. The figure describes the ratio of cortical columns in each rung during periods of 5000 seconds when L6a was in the *Known* state. This is denoted as ratio of long-term cortical stability. The cortex is considered stable if all CCs are long-term cortically stable, indicating no CA3 allocation.

### E. Acceleration of Identification (Corticothalamoidal Loop)

After evaluating the positive effects of the proposed learning optimizations, we will now investigate how the pattern matching acceleration works with the current system. We propose adding CTLoop on top of the already analyzed mechanism. We conservatively set the matching requirement to be perfect, which means that the result of masking L5 with L6+ should have the same size as the actual values expected. Four replicas of layer L5 per module in the CCs of the third rung are used. The CC in the fourth rung does not have an L5 layer. The results are shown in Figure 20.

Although requiring an additional layer (L5), according to Figure 20.A, the value of synaptic load in upper rungs is slightly lower than that observed without CTLoop. Similarly, the time to reach steady state is shorter. This is because the CTLoop not only accelerates but also stabilizes, resulting in improved performance. In this scenario, Perfect EOS is not available, since L5 recurrent predictions cannot be tracked, hence we use long-term cortical stability to assess cortical knowledge. According Figure 20.C when the simulations end, all CC are stable in the long-term.

Despite the complexity of the problem, as can be seen in Figure 20.B, on average, the third rung is able to forward nearly a 40%[13] of the output. This translates to a reduction in the sequence length of the third rung around 30%. Therefore, the system can identify the sequences in the top rung of the hierarchy faster. Although tweaking the parameters of the loop (e.g., increasing the matching size) or adding the missing corticostriatal loops may improve the figures, it is still a remarkable result. Not only does the idea increase the speed of identification, but it also makes the system less sensitive to a significant portion of sensory input. In contrast to the top rung, correct predictions in the lower rungs are quite rare. The reason for this is that it requires not only downstream attention (i.e., only half of the time at most) but also preventing misalignment in lateral contextualization. The two CCs attached to the CA3 above enable it. On average, this will limit the forwarding window to 25% of the time in the second rung and 12% in the first rung. Given that the predictions on those rungs are wider, it follows that the forwarding is infrequent. As discussed at the end of Section V.B this is not a limitation of the idea. The risk-benefit is optimal in the top rung, and the lack of forwarding on the lower rungs allows for easy reconnection of missed predictions in the top rung.

---

[12] Note that the low number of synapses per branch is a consequence of the RNSC principle: only a few cells own the majority of actively used branches. These results consider the fact that first rung CC has only one module, and that the third and fourth rungs haves two and four out scale replicas.

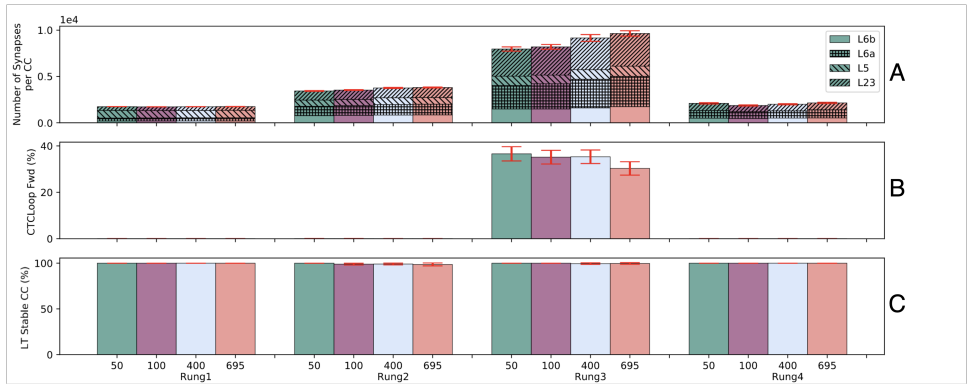

Figure 20 Results for sequentially fed *Sharvard* corpus with Corticothalamoidal Loop, A: Synaptic Load, B: Percentage of output predicted, C: Long-Term Cortical Stability

Looking at other metrics, it appears that both synaptic load and sequence length also improve slightly. Although CTLoop is a mechanism for improving identification speed, it also introduces additional stability into the system, which has a positive impact on both. When CTLoop is introduced into the system, it is not possible to rely on the Perfect EOS ratio as a metric of cortical stability, since predicted events are untraceable. In this situation, we use an alternate metric, which is the percentage of CC that is fully stable (i.e., with no CA3 allocated above or below). In non-CTLoop systems, this metric is ahead of Perfect EOS, but when 100% is reached, Perfect EOS starts to converge to the final value fairly quickly.

Those results can explain event-related potential (ERP) in the electroencephalogram (EEG), including language-related effects like N400 and P600 [173]. In these events cognitive effects, like other ERP, are reflected in changes in the EEG. For example, N400 is a phenomenon that occurs when an inconsistent word (i.e., a semantic anomaly) is presented within a sentence. This leads to a decrease in the voltage detected by the EEG electrodes in specific areas of the scalp, with a delay of 200-400ms following the event's onset. The discovery of the link between cognitive and electrophysiological effects almost 25 years ago generated significant interest due to its potential. However, current neuroscience still struggles to determine its functional relevance [174]. Our hypothesis is that CTLoop may shed light on the conundrum. When CTLoop is functioning properly, the last rung will operate in burst mode, with short periods of high activity (such as the *S* sequence in Figure 8) followed by long periods of silence. A relatively large rung of the cortex (e.g., using the out-scale mechanism depicted in Section V.E) will be most likely detectable by EEG electrodes. If there is a prediction mismatch burst activity will vanish which may appear as a lack of activity in the EEG. Based on our findings, the lower rungs are more likely to function with actual input, resulting in faster reconnection with the real input compared to the length of the unexpected word, and a reduction in EEG activity. P600 may operate in the opposite direction. When the listener becomes aware of the sentence's meaning (i.e., predictions begin to align), there will be an increase in activity due to the burst of action. The 200-500ms delay might be related with the fact that the third rung is operating during half of the sequence length. In the absence of CTLoop, the average sequence length is approximately 1.2 seconds with 50% silence in the input and a 40% hit rate in predictions, resulting in a delay of approximately 360ms (for our experimental setup). It will take approximately 360ms to reconnect the last rung with the real data. P600, which is an increase in EEG activity following an unexpected syntactic structure, may be due to the opposite effect. After a difficult-to-process syntactic structure, the fourth rung begins to adjust to the predictions from the lower rung. The delay may be related to the end of the unexpected word and the beginning of the first predicted word. Based on this discussion, CTLoop can not only explain other ERP effects too, but also support the highly speculative hypothesis presented in Section V.B. regarding the origin of the EEG signals.

---

[13] The simulations were stopped when the simulated clock reached ~2 million seconds. These figures can be improved slightly by keeping the system running for longer.

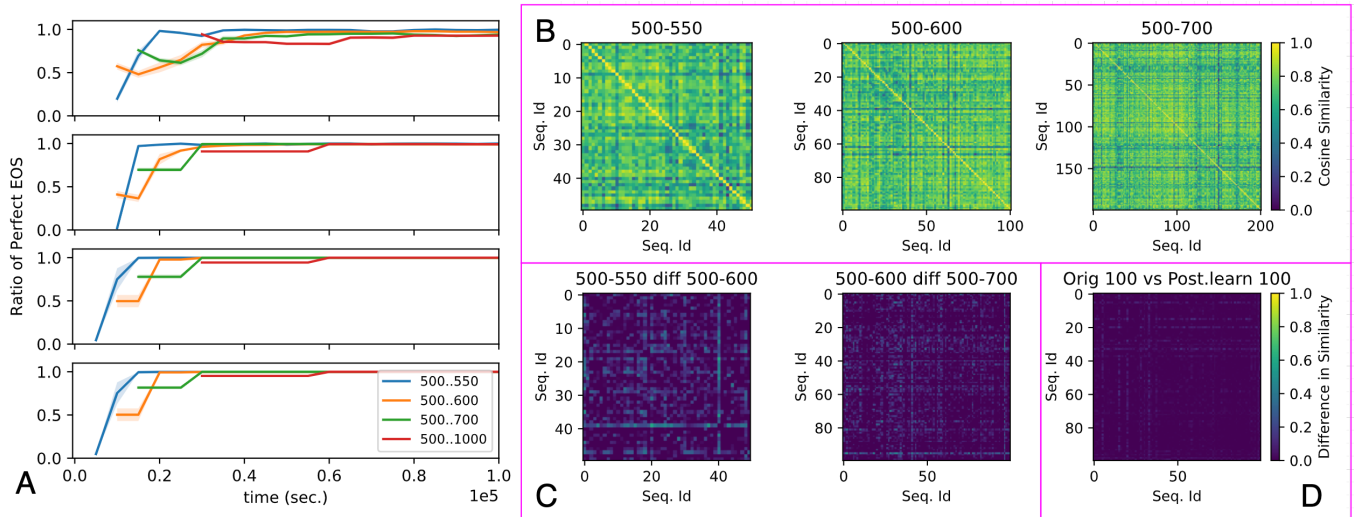

**Figure 21** Compositional Learning, A: Temporal Evolution of Perfect EOS ratio, B: Cosine Similarity of cortical output with 50, 100 and 200 new sentences, C: Change in cosines of shared sentences through datasets, D: Change in cosine similarity of the encoding of the first 100 sequentially exposed sequences before and after learning 500 new sentences.

## IX. RESULTS II: PUTTING IT ALL TOGETHER

One may question the usefulness of the proposal: is it merely a memory system could it learn and seamlessly identify and/or incorporate new information into its internal model under realistic usage scenarios? To address this question, we will now shift our focus to a more representative, and demanding usage scenario: realistic speech [160]. The cortex is fully configured (i.e., includes BG, SWR and CTLoop), and uses 12 audio features (i.e., Cortex4 in Figure 13).

### A. Compositional Learning
### (and Catastrophic Learning avoidance)

To start, the initial 500 sentences of the dataset are fed into an empty cortex using random sorting. In contrast to *Shavard*, these sentences are written in a more natural style, as they are taken from acclaimed works of Spanish literature. Multiple simulations are run until all cortical columns reach a stable state. The simulation with the synaptic load closest to the average is selected. Multiple simulations with the first 100 sentences in sequential order are run until the percentage of perfect EOS[14] reaches at least 98% in all rungs. The chosen cortical state is the one closest to the average among all simulations.

This state is loaded into a cortex and fed sequentially with the next 50, 100, 200 and 500 sentences, starting with the last sentence (i.e., from 500 to 550, 500 to 600, 500 to 700, and 500 to 1000). To ensure statistically significant results, each input is run multiple times until the percentage of EOS reaches at least 98% in all rungs. Figure 21.A presents the temporal evolution of the average ratio of perfect EOS for each input. The system acquires new knowledge very quickly, as shown in contrast to Figure 19. Even with more complex sentences, the system reaches stable behavior for 50 sentences in approximately 25,000 seconds, which is about three orders of magnitude faster than when learning from scratch. Moreover, the time required is sublinear with the size of the new information. The synaptic load of adding this new information is negligible, falling within the statistical confidence margins. Another relevant fact is that the dispersion from sample to sample in behavior is limited. This suggests that stochasticity impact on the learning outcome is minor. Note that initial connectivity is the same. Hence, in subsequent experiments multiple simulations per data point are unnecessary.

Although interpreting cortical output can be complex, we analyzed how the provided system representations compare and how the new knowledge distorts the original. We use the time-agnostic encoding discussed in Section VIII.A. For each tag change in any CC of the last rung of the first simulation, we annotate the codes to build a multidimensional representation of the previous one. We compute the cosine similarity among the sequences, which is shown in Figure 21.B for the first three experiments. Although time interleaving is not considered, there is a clear diversity of representations that indicates the practical usefulness of HER. Figure 21.C shows how each experiment's portion affects the subsequent one by providing the difference between the similarity matrices. The matrices are very similar, indicating that the original cortical state conditions the learning outcome, which corroborates the aforementioned low dispersion in high level metrics.

---

[14] To use this metric, it is necessary to remove CTLoop in the cortex because forwarded predictions may distort it.

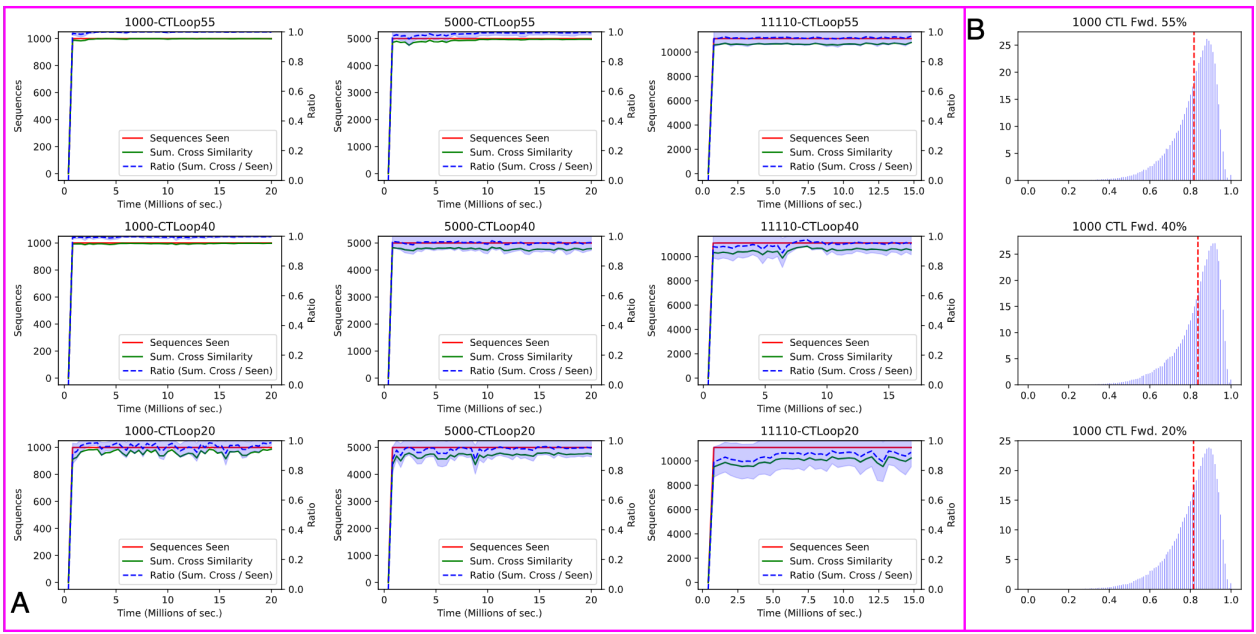

**Figure 22 A:** Representational drift for different inputs for 55%, 40% and 20% CTLoop,
**B:** Histograms of Similarity of cosine for 1000 sequences with different CTLoop forwarding.

Finally, one may wonder if newly acquired knowledge compromises previously acquired knowledge. Figure 21.D shows the difference between the similarity matrices of the first 100 sentences before and after learning an additional 500 sentences. The cosine similarity of the matrices is 0.9991, suggesting that catastrophic forgetting [175] is not present in the system. The average drift in the encoding is below 2.5º, which may fall within the natural behavior of the system[15].

From these results it appears that, in contrast to DL systems [176], as biological systems [177], HER is able to perform compositional learning seamlessly. However, it is important to note that these results do not necessarily indicate that the system "comprehends" any data. Rather, they suggest that compositionality of language may be achievable. Although this result may seem unforeseen at first, it is not so surprising when considering the true hierarchical nature of HER[16]. The concept of compositionality, while elusive and controversial, may be an emergent property of hierarchical systems rather than the other way around (i.e., designing the system based on the compositionality of the behavior to mimic [178]). Although large language models appear to have composability capabilities [179], these are based on prompt engineering [180], rather than changes to the model itself. In contrast, our proposal incorporates newly learned information into the model. The inherent properties of Stochastic Gradient Descent prevent Deep Learning from achieving this crucial feature [175].

### B. Continuous Learning Support (CTLoop in a Naturalistic Environement)

The previous experiment has shown that the system can acquire new knowledge more quickly when starting with a foundational model, without affecting what is already known. However, it is important to clarify the system stability when exposed to a significantly larger amount of information over extended periods of time. To increase system stability with continuous learning, it is necessary to integrate the internal predictions of the cortex into the sensory stream. The proposed

---

[15] In contrast to DL systems, the output representation is not fully deterministic. There is always a representational drift due to the continuous learning.

[16] Although DL approaches appear to be hierarchical in structure, the back-propagation algorithm makes it unclear whether weights in the early layers are reused between inputs, as in HER. For this reason, unless there are specific mechanisms, such as[193]. SGD suffers from catastrophic forgetting. Other non-SGD, such as continuous learning network architectures, also suffer from the same problem [194].

CTLoop mechanism is fundamental for this task. It allows the system to ignore the sensory stream that falls within the forward predictions, which is not relevant from a reliability perspective, as previously discussed. This is relevant not only for ensuring reliability, as previously discussed, but also for constraining representational drift caused by continuous learning.

Then we reinstate the use of CTLoop to have a fully functional cortex. To quantify the representational drift of each sequence we annotate the average code of each one each 400.000 seconds and compute the cross-cosine similarity with the previous episode. The ratio of the sum of all the cross-cosine similarities and the total number of symbols in time provides a view of how the system is reacting in time. Figure 22.A introduces these results using three different foundational states with increasingly larger number of new sentences. Each foundational state is selected from a set of sixteen simulations of an empty cortex learning 500 randomly sorted sequences until all CC are stable. Although on average the forwarding probability is 40%, there is spectrum of cases from 55% to 20%. We chose three simulations in both ends and in the middle. These three cortices are then exposed to a growing number of sentences never seen during the foundation learning, with a growing portion of the dataset (from 1000, 5000 to the whole data set). In each case the system is exposed sequentially to up to 20M seconds of continuous audio (~0.63 years). In all cases the system identifies the complete set of sequences However, there is a noticeable change in the ratio of the sums of the cross-cosine similarity and the number of sequences, as we increase the number of sequences. This ratio, represented as a dotted line, remains above 0.99 for 40% and 55%, and starts to degrade as we increase the input, especially with CTLoop with just 20% forwarding. Such ratio, represented as a dotted line remains above 0.95 in all cases for 1000 sentences but stars to degrade as we increase the input, especially whit 20% CTLoop. In this case, when the average falls below 0.9, the change in the vector is approximately 25º. A higher CTLoop forwarding not only reduces the average ratio but also decreases the variation, as shown by the narrower blue shaded area. This behavior is consistent with the intuition that a higher degree of forwarding may increase the system capacity.

One may question whether the stabilization provided by CTLoop forwarding comes at the cost of reduced disambiguation capabilities. As shown in Figure 22.B, for 1000 sentences, where the representational drift is similar, the histogram of the cosine similarity matrix is close. This indicates that CTLoop does not negatively affect the cortex's disambiguation capability.

The results indicate that CTLoop must be enacted frequently to identify and learn patterns in a naturalistic environment. Our current model, which lacks a significant portion of the corticostriatal loop found in biological systems, may be limited in achieving such accuracy. Future work should focus on developing the missing piece of the algorithm. Nonetheless, the current proposal can still be utilized by selecting the model with the highest accuracy from a large pool of simulations, as it has been done here.

X. RELATED WORK

Today, most techniques in ML are closed-ended, where an objective function is pursued via non-biologically plausible optimization processes. HER uses an open-ended approach [181]: the system is reorganized according to input exposure without any explicit objective. Hence, none of these techniques are comparable to HER. Note that the algorithm design has embedded the objective within minimizimg the synaptic load without losing accuracy in the predictions. From an evolutionary standpoint, this is equivalent to maximizing the reach of the internal model of the world, given a constrained set of resources (i.e., some form of sophisticated compression). Maximizing prediction accuracy and reach is advantageous for survival, which might be encoded in the system configuration as innate information [164]. This is very distant from conventional ML approaches. The most similar work to this proposal, to the point of being the main inspiration of it, is the HTM ideas developed by J. Hawkins and colleagues [20]. Like our work, the original authors have evolved these HTM into the definition of a plausible organization for the cortex [182]. Following a top-down reasoning, a reference frame-based approach is proposed, called Thousands of Brains Theory (TBT). The idea is mainly based on the location specificity exhibited by some cells in the hippocampus complex region (place-cells, grid-cells, head-cells, etc…) and their hypothetical presence through the cortex. In contrast with our work, centered in achieving a practical system, the objective of TBT is to develop a plausible hypothesis about the cortical column functionality. Contrarily to the commonly held hierarchical view of the cortex, the key concept is that every part of the cortex learns complete models of objects. The non-hierarchical interaction between columns enables, via voting, to resolve sensory input ambiguity.

Our work, following a bottom-up approach, tries to achieve certain design objectives for an artificial system. Maintaining biological plausibility, we speculatively hypothesize about of how the cortex might work. In any case, this is not the drive of the proposal. Our proposal development is based on the learning process (i.e., how to acquire a model of the world at bootstrap) and not so much on how that model must be interpreted from a high-level perspective. In our proposal a fundamental design goal is to minimize the resources (i.e., synaptic load of the system and energy expenditure) without affecting the disambiguation capabilities (i.e., from the sensory information distill a minimal, yet useful, model). This conflicts with the notion of multiple replicas of the modeled objects across the cortex of TBT. In our case, the system proposed is aligned with the classical hierarchical view of the cortex, in our view, is the most reasonable approach, for a given world model complexity, to minimize synaptic load. In our work we use near replication to compensate for the lack of disambiguation of high dimensionality reduction (which is required to achieve invariance and increase system robustness). The replication occurs within the cortical column not across them. Most of the connectivity between columns has short range. Only relatively few long connections are required for lateral contextualization. Note that although in our case CTLoop can be used to identify a whole object with a few features (in the appropriate context), we do not require breaking the hierarchical organization.

We progressively build a model that suggests how other elements, auxiliary to the cortex, such as the thalamus, hippocampus, basal ganglia and brainstem, might play a critical role in its learning and operation. In TBT such elements are not discussed in detail how might or might not impact the viability of the model. In our model, we explain how some are helpful (CTLoop) and some are critical (HC).

In contrast with TBT, the reference framework used by each cortical column here is not location but time (via L6b sequence segmentation). Given that our work is mainly focused on auditory signals, movement and/or embodiment does not seem necessary, while subcortical projections of the cortex are used as the reference frame. Additionally, as discussed before, in an action-dependent sensory stream, such as vision or touch, might also fit within a temporal-based framework. Our proposal suggests that certain brainstem pathways (and the interaction with the cortex expectations) have a critical role in order to transform the location-based sensory information into a stable temporal stream of data, whereas in TBT that role is not discussed for subcortical structures. Hence, in contrast to TBT, which assumes that the cortex is handling location-based information by building up reference frames, our model assumes that the cortex is only perceiving stable temporal information. Consequently, the hypothesized role of the hippocampal formation in both works is dissimilar.

The transitory integration of feedback and cortex expectations in TBT is not discussed. Similarly, to Predictive Coding, one might argue about how reference input (such as motor) is learnt in first place. When the cortex expectations appear, to suddenly modifying the sensory input might introduce instabilities. HER contemplates how to seamlessly align the cortex knowledge with the input signal (and rungs in the hierarchy) across learning. In fact, feedback is not considered in any form in TBT, whereas in our algorithm it is a key element at many levels.

Our proposal is loosely related with Adaptive Resonance Theory (ART) [183]. Its core principle is to adjust system state to achieve synchronized operation between the input stream and previous experience by inducing periodic resets in divergences (called the cycle of reset and resonance). This is like our sequence boundary events. Additionally, our "*Lateral Contextualization*" might resemble the complementary attention proposed by ART. ART fundamentally faces the "*Stability-plasticity dilemma*". This is also present in our proposal and addressed (like in ART, locally) by the learning hysteresis cycle. Nevertheless, similarly to TBT, ART follows a top-down approach (even, from a cognitive standpoint, it starts at a higher level). In contrast with our work, which uses simulation as the main exploration tool, ART is strongly founded in a complex theoretical corpus (in part due to the lack of computational power during its inception and evolution). ART's objective is to close the gap between cognitive phenomena and experimental evidence. Nevertheless, some current neuroscience knowledge, such as the post-synaptic effect of feedback synapses, seems to be against some of the basic assumptions of ART. In any case, the models hypothesized for the cortical column or other support structures in ART are very different to ours, both in the core principles and in operation mode. In our case, the critical features on the data stream appear (at any level) due to frequent occurrence and are not actively searched for by design. In addition to the above examples, there are a significant number of so-called cognitive architectures. While some, like ART, are more focused on physiological or agent plausibility (e.g., SOAR[184]), others are based on neuroscience plausibility (e.g., SPAUN[185], Leabra[186]). Our proposal, thanks to the level of abstraction chosen and despite being based on the latter' plausibility, has numerous connections with the former.

From the perspective of a complementary level of analysis [187], previous work, such as the Jeffress sound localization model [188], has followed a methodology similar to ours. The model was confirmed much later by biological evidence [189]. Although not a primary goal of our work, the chances of not being wrong may be no zero. We believe that top-down efforts often rely on subjective intuitions, which may be distant from the underlying reality. Building on the ideas presented in [6], it appears highly challenging to comprehend or reproduce the internal mechanisms of a contemporary computer solely through the lens of the user interface. Therefore, it seems that a middle-up approach would be the most effective means of achieving the ultimate objective.

XI. CONCLUSIONS

The results obtained seem to suggest that the proposed approach might be useful in practical problems. Certainly, to be competitive with state-of-the-art machine-learning requires an engineering effort that cannot be committed at this point in time. In any case, to the best of our knowledge, there is no other competitive proposal able to perform continuous unsupervised learning with such resilience.

From a high-level perspective, our proposal fits more within associative learning theory than an information-processing framework [190]. Hence, domain-specific learning mechanisms are unnecessary [191]. Consequently, the extension to other domains is feasible. Although, according to the stimuli's nature, some input-dependent adaptation must be done, the algorithm of the back end will remain unaltered.

Beyond potential domain extensions, this work could be used as a first order approximation to be falsified (or not) by neuroscience. Many guesses, such as astrocytic processes calcium variations, thalamo-cortical pathway, hippocampal circuit, etc… can be considered. As [192] suggests, the virtuous circle between machine learning and neuroscience could be a potential path to progress in one of the most challenging problems in science. The computer architecture point of view (which is weakly considered by most ML practitioners) can steer what machine-learning mechanisms can be reconciled with the power and reliability constraints of a physical system. We feel that the synergy of the three fields can solve the issue. This work is just an early example of it.

ACKNOWLEDGMENTS

To José Angel Gregorio for those long, heated, and productive discussions. Without them this work could not have taken place.

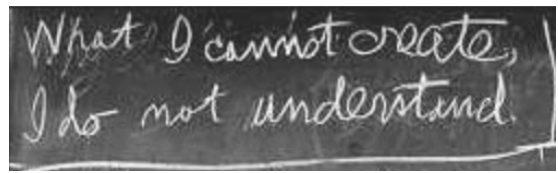